\documentclass[letterpaper, 10 pt, conference]{ieeeconf}  

\IEEEoverridecommandlockouts                              

\usepackage{amsmath} 
\usepackage{amssymb}  
\usepackage{booktabs}
\usepackage{graphicx}
\usepackage{url}
\usepackage{xurl}
\usepackage{adjustbox}
\usepackage{balance}
\usepackage{algorithm}
\usepackage{algpseudocode}
\usepackage{tcolorbox}
\usepackage{booktabs}
\usepackage{multirow}
\usepackage{listings}
\usepackage{placeins}
\usepackage{capt-of}
\usepackage{booktabs}
\usepackage{tabularx}
\usepackage{array}
\usepackage[hidelinks,bookmarks=false]{hyperref}

\lstdefinestyle{prompt}{
    basicstyle=\ttfamily\footnotesize,
    breaklines=true,
    breakatwhitespace=false,
    columns=fullflexible,
    keepspaces=true,
    showstringspaces=false,
    frame=single,
    literate={→}{{$\rightarrow$}}1
}

\title{\LARGE \bf
R2S-Eval: Robot Evaluation with Real-to-Sim Calibration via Vision-Language Models
}

\author{
Yidi Wang$^{1,2,*}$,
Feixiang Ruan$^{1,3,*}$,
Ruoqu Chen$^{4}$,
Jie Yin$^{1}$,
Yang Yu$^{2}$,
Mengdi Xu$^{4}$,
Kaifeng Zhang$^{1,\dagger}$
\thanks{$^{*}$Equal contribution; $^{\dagger}$Corresponding author; $^{1}$Sharpa; $^{2}$Nanjing University; $^{3}$Tongji University; $^{4}$Tsinghua University}}

\begin{document}

\maketitle
\thispagestyle{empty}
\pagestyle{empty}

\begin{abstract}

Evaluating robot manipulation policies is becoming increasingly important as generalist models, particularly vision-language-action (VLA) models, are deployed on physical robots. However, conventional real-world evaluation remains labor-intensive, unstable, and insufficiently informative. It requires repeated hardware trials, manual scene resets, and continuous operator monitoring, may produce different policy rankings across repeated evaluations, and primarily relies on success-rate metrics that provide limited information about execution quality. In contrast, humans assess robot performance by observing and comparing complete behaviors rather than relying solely on binary success outcomes. To this end, we propose R2S-Eval, an evaluation pipeline that combines real-to-sim calibration with vision-language model (VLM) preference evaluation. The real-to-sim component efficiently generates rollout videos in a simulator
calibrated to the real-world evaluation setting, thereby reducing the need for repeated hardware trials. The VLM evaluator assesses the execution quality of rollout videos and produces pairwise preferences, which are subsequently aggregated into policy rankings. We further introduce a protocol to assess whether the proposed evaluation pipeline yields validated policy conclusions while mitigating the key challenges of conventional real-world evaluation. Experiments in both simulation and real-world settings demonstrate that R2S-Eval produces reliable and stable policy conclusions, achieves agreement with human preferences, substantially reduces repeated hardware-operation effort, and reveals behavior-quality differences that are not captured by binary success labels. In general, R2S-Eval advances robot evaluation from manual success counting toward automated, statistically stable, and quality-aware evaluation of robot behavior. Project page: \url{https://r2s-eval.github.io}.

\end{abstract}

\section{Introduction}

\begin{figure*}[t]
    \centering
    \includegraphics[width=0.95\textwidth]{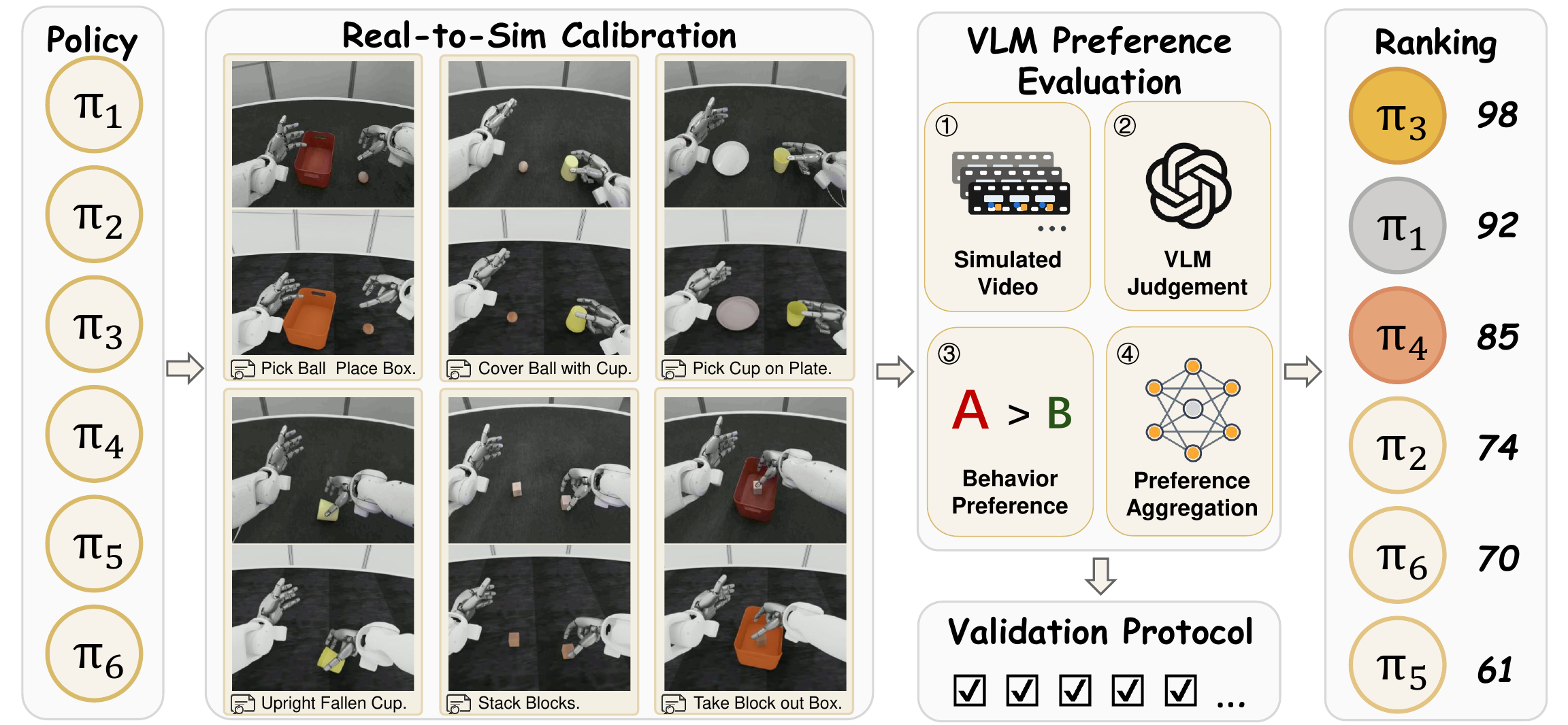}
    \vspace{-0.5em}
    \caption{Overview of the R2S-Eval pipeline for ranking robot policies using real-to-sim calibration and VLM-based preference evaluation.}
    \label{fig:overview}
\vspace{-0.5em}
\end{figure*}

Recent progress in generalist manipulation policies, such as
vision-language-action (VLA) models, is bringing physical robots closer to solving real-world tasks~\cite{
brohan2022rt,
zitkovich2023rt,
driess2023palm,
kim2024openvla}.
In this context, evaluation is essential for measuring how well different policies perform in real settings. A well-designed evaluation does more than report performance. It reveals strengths and failure modes, guides data collection and training, and helps identify how to improve robot policies.

Current policy evaluation typically follows a protocol of task-specific fine-tuning and hardware deployment. Candidate policies are fine-tuned on downstream task data and deployed in a manually configured real-world environment, where operators conduct repeated trials and record task successes. Policies are ranked by their success rates, and the resulting ranking is treated as a proxy for the real-world manipulation capability of the underlying generalist models~\cite{
kim2024openvla,
liu2023liberobenchmarkingknowledgetransfer,
li2024evaluating}.

Although effective, this hardware evaluation loop is labor-intensive and unstable. Each trial requires scene setup, robot execution, object reset, and hardware monitoring, leading to a growing manual effort. The resulting rankings are difficult to repeat consistently, as small changes in object placement, contact, perception, or robot state can alter rollout outcomes~\cite{li2024evaluating}. Moreover, the success rate is too coarse and provides limited information on execution quality. Policies with similar success rates may differ substantially in motion quality, corrective behavior, or progress before failure.

These limitations motivate a more informative view of robot policy evaluation, one that is closer to how humans assess robot behavior. When watching a rollout, humans consider the complete execution rather than only whether the final goal is reached. They can distinguish two successful rollouts that differ in control quality, or two failed rollouts that make different amounts of progress toward the goal. This observation suggests evaluating policies from rollout videos and estimating policy quality through preferences over observed executions rather than only success counts~\cite{
christiano2017deep}.

However, video-based preference evaluation is not by itself practical. Collecting rollout videos on hardware still requires repeated manual effort, while video comparison introduces additional annotation effort. A practical video-based preference evaluation must therefore address two problems: how to efficiently obtain behavior videos consistent with the real world, and how to judge behavior quality automatically in a way that reflects human preferences.

To this end, we propose R2S-Eval, illustrated in Fig.~\ref{fig:overview}, an evaluation pipeline that combines real-to-sim calibration with vision-language model (VLM) preference evaluation. The real-to-sim component constructs a simulation calibrated to the real world. Candidate policies are adapted with simulated data and deployed in closed loop in simulation to produce rollout videos efficiently. A VLM then assesses the execution quality of rollout videos and compares sampled video pairs to produce preferences that are aggregated into policy rankings. We further introduce a validation protocol that examines whether these resulting conclusions reflect model capability, stabilize with increasing trials, agree with human preferences, reduce repeated hardware-operation effort, and reveal behavior differences beyond success counts.

In summary, the main contributions of this work are:
\begin{itemize}
    \item We formulate manipulation policy evaluation as preference estimation over rollout behaviors, producing policy rankings that reflect execution quality rather than binary success counts.

    \item We introduce R2S-Eval, an evaluation pipeline that obtains behavior videos from a calibrated real-to-sim simulation and ranks policies using VLM-generated preferences.
    
    \item We provide a multi-axis validation protocol for policy evaluation methods covering ranking reliability and stability, human agreement, hardware manual effort, and behavioral informativeness beyond success.
    
\end{itemize}

\begin{figure*}[t]
    \centering
    \includegraphics[width=0.95\textwidth]{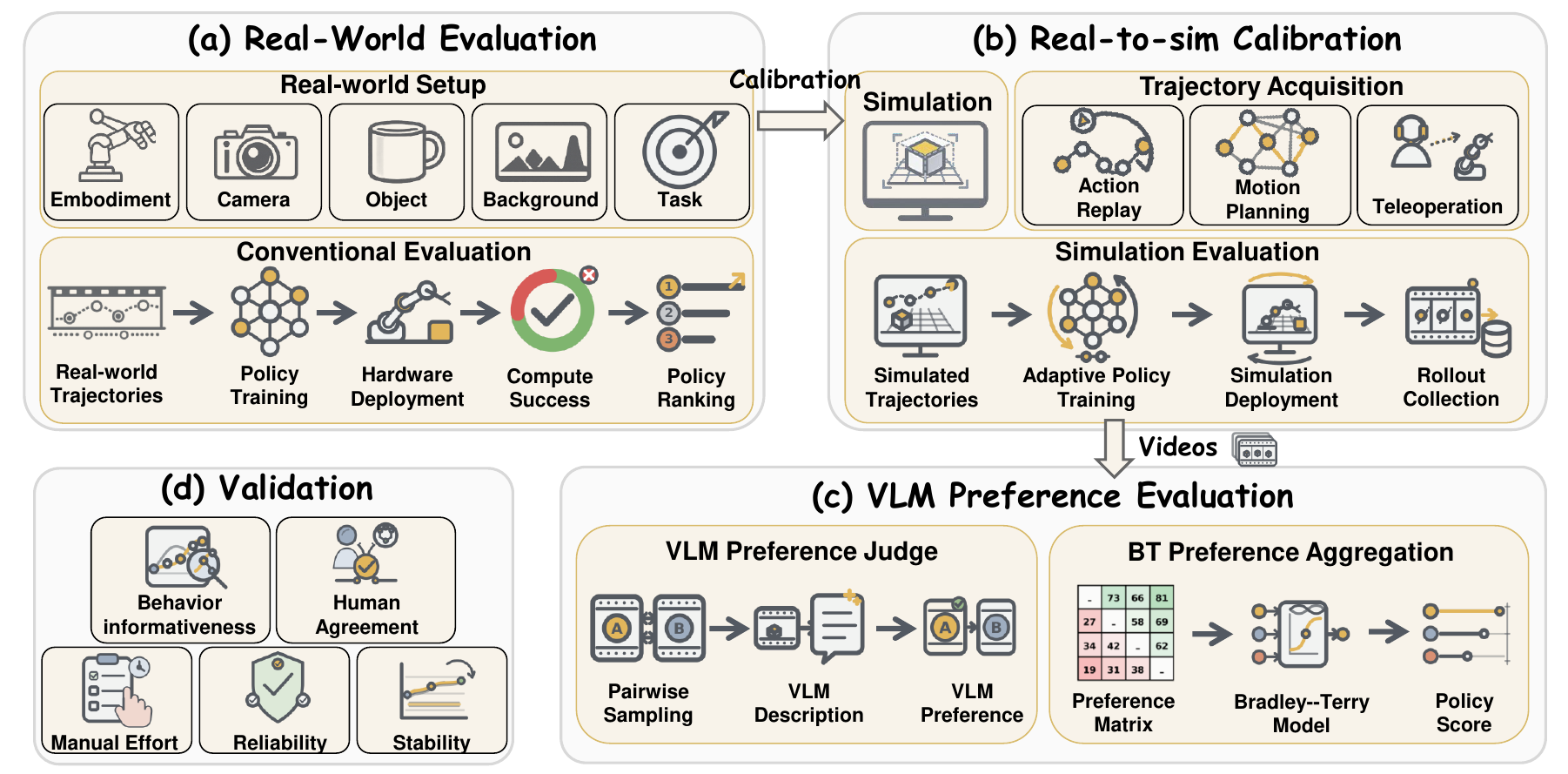}
    \vspace{-0.5em}
    \caption{Detailed workflow of the R2S-Eval pipeline. To address the limitations of (a) real-world robot evaluation, (b) the calibrated real-to-sim simulator generates rollout videos for (c) VLM-based preference evaluation. (d) The full pipeline is validated.}
    \label{fig:method}
\vspace{-0.5em}
\end{figure*}

\section{Related Work}
\noindent\textbf{Real World and Simulation Policy Evaluation.}
Robot policy evaluation spans standardized simulation benchmarks, automated real-world testing, and proxy evaluation through simulators or learned world models. Recent benchmarks assess language-conditioned manipulation, long-horizon reasoning, robustness, and cross-task generalization~\cite{wang2023robogen,james2020rlbench,zhu2020robosuite,yu2020meta,gu2023maniskill2,liu2023liberobenchmarkingknowledgetransfer,zhang2024vlabench,chen2025robotwin20scalabledata}. Real-world systems automate success detection, reset, and policy scheduling, or distribute blinded comparisons across sites~\cite{zhou2025autoeval,atreya2025roboarena}, while proxy methods study whether simulated performance predicts hardware results~\cite{li2024evaluating,li2025worldeval}. However, these approaches still emphasize success rates, scalar scores, or aggregate rank correlations, leaving behavioral differences and ranking stability insufficiently characterized. Large-scale embodied benchmarks such as BEHAVIOR-1K~\cite{li2024behavior} further expand evaluation toward long-horizon household activities, but evaluation is still primarily based on task completion metrics.

\noindent\textbf{Real-to-Sim Environment Construction.}
Real-to-sim methods reconstruct physical environments for policy training and evaluation using geometric reconstruction, neural rendering, 3D Gaussian Splatting, and physical simulation~\cite{robocasa2024,kerbl20233dgaussiansplattingrealtime,torne2024reconcilingrealitysimulation,li2024robogsim,han2026re3sim,zhang2025realtosimrobotpolicyevaluation}. Related work transfers video-reconstructed hand-object interactions into simulation to recover tactile supervision~\cite{chen2026dexx}. Recent frameworks further use reconstructed environments for sim--real comparison, progress estimation, and preference aggregation~\cite{jain2025polaris,jangir2025robotarenainfty}. Nevertheless, contact, detection, and control mismatches remain difficult to reproduce, and evaluation is still dominated by task results or aggregate agreement.

\noindent\textbf{VLM-Based Robot Behavior Evaluation.}
VLMs have been used for success detection, task-progress estimation, failure diagnosis, subgoal verification, and execution-quality assessment~\cite{du2023visionlanguagemodelssuccessdetectors,duan2024aha,liu2026evalactions}. Although these methods capture process-level differences beyond binary success, most assign absolute labels or scores to individual trajectories. Existing policy ranking approaches also rely heavily on expert annotations or learned evaluators~\cite{liu2026evalactions}. Recent LLM-as-a-Judge studies demonstrate the effectiveness of pairwise preference evaluation for ranking complex model outputs~
\cite{
zheng2023judging,
chiang2024chatbot}. Aggregating repeated pairwise VLM judgments into statistically grounded and validated policy rankings remains underexplored.

\section{R2S-Eval Pipeline}

In this section, we present the R2S-Eval evaluation pipeline. Sec.~\ref{formulation} formulates the evaluation of policies as ranking candidate policies based on preferences over rollout behaviors. Sec.~\ref{real2sim} describes real-to-sim calibration, collecting closed-loop rollout videos in a real-world calibrated simulation. Sec.~\ref{vlm_eval} introduces the VLM preference evaluation, where a VLM judges videos and generates preferences that are aggregated into policy rankings. Sec.~\ref{validation} presents a validation protocol to examine the evaluation pipeline.

\subsection{Problem Formulation}
\label{formulation}

We consider the problem of evaluating a set of candidate manipulation policies
\(\Pi=\{\pi_1,\pi_2,\ldots,\pi_N\}\) under a target evaluation distribution
\(\mathcal{D}\), which specifies the tasks, initializations, and environment settings. A rollout of policy \(\pi_i\) is
denoted by \(\tau_i^k=(o_0,a_0,o_1,a_1,\ldots,o_T)\), where \(o_t\) and
\(a_t\) are observation and action at time \(t\). We denote the video of a rollout by
\(v_i^k=\phi(\tau_i^k)\). The goal is to produce a ranking
\(r\) over \(\Pi\), where higher-ranked policies are judged to have better
manipulation capabilities under \(\mathcal{D}\).

Conventional evaluation, shown in Fig.~\ref{fig:method} (a), ranks policies according to task success. Let
\(c(\tau)\in\{0,1\}\) indicate whether a rollout is successful. The ranking is obtained by sorting the empirical success rate:
\begin{equation}
\begin{aligned}
    \hat{s}^{\mathrm{succ}}_i
    &=
    \frac{1}{K_i}\sum_{k=1}^{K_i} c(\tau_i^k), \\
    r^{\mathrm{succ}}
    &=
    \operatorname{Rank}
    \left(
    \hat{s}^{\mathrm{succ}}_1,\ldots,
    \hat{s}^{\mathrm{succ}}_N
    \right),
\end{aligned}
\end{equation}
where \(\operatorname{Rank}(\cdot)\) orders policies. Since \(c(\tau)\) reduces each rollout to a binary outcome,
\(r^{\mathrm{succ}}\) ignores the behavior differences among rollouts with the same success label. 

We instead formulate the policy evaluation as preference estimation
over rollout behaviors. Let
\(\mathcal{V}_i=\{v_i^1,\ldots,v_i^{K_i}\}\) denote the rollout videos of \(\pi_i\).
For a sampled pair \((v_i^a,v_j^b)\), a preference judge outputs
\(y_{ij}^{ab}\in\{i\succ j,\;j\succ i\}\). Let \(\mathcal{C}_{ij}\) denote the comparisons
between \(\pi_i\) and \(\pi_j\). The observed preferences are
summarized by a matrix \(M\in\mathbb{N}^{N\times N}\),
\begin{equation}
    M_{ij}
    =
    \sum_{(a,b)\in\mathcal{C}_{ij}}
    \mathbf{1}\!\left[y_{ij}^{ab}=i\succ j\right]
\end{equation}
where \(M_{ij}\) counts how often rollouts from \(\pi_i\) are preferred over
rollouts from \(\pi_j\).

To infer a ranking from \(M\), we introduce a statistical preference
model. Let \(\theta_i\) denote the latent quality score of the policy \(\pi_i\) and \(q_{ij}(\theta)=\Pr_{\theta}(\pi_i \succ \pi_j)\) be the probability that \(\pi_i\) is preferred over \(\pi_j\). For independent binary comparisons, the
log-likelihood of \(M\) is
\begin{equation}
    \ell(\theta;M)
    =
    \sum_{i<j}
    \Big[
    M_{ij}\log q_{ij}(\theta)
    +
    M_{ji}\log q_{ji}(\theta)
    \Big],
\end{equation}
where \(q_{ji}(\theta)=1-q_{ij}(\theta)\). Then the scores and ranking are
\begin{equation}
\begin{aligned}
    \hat{\theta}
    &=
    \arg\max_{\theta}\ell(\theta;M),\\
    r^{\mathrm{pref}}
    &=
    \operatorname{Rank}
    \left(
    \hat{\theta}_1,\ldots,\hat{\theta}_N
    \right).
\end{aligned}
\end{equation}
Thus, policy evaluation shifts from sorting success
counts to estimating a ranking from preferences over
rollout behaviors.

\begin{figure*}[t]
    \centering
    \includegraphics[width=1.0\linewidth]{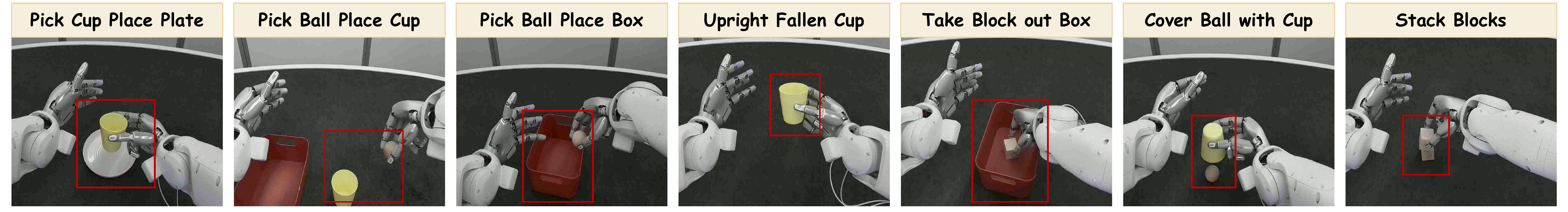}
    \vspace{-0.5em}
    \caption{Illustration of the real-world manipulation task suite in experiments.}
    \label{fig:tasks}
\vspace{-0.5em}
\end{figure*}

\subsection{Real-to-Sim Calibration}
\label{real2sim}

Sec.~\ref{formulation} assumes access to the rollout videos
\(\mathcal{V}_i\) for each candidate policy. Directly collecting these videos inherits the limitations of conventional real-world evaluation, which is labor-intensive and unstable. To address the video acquisition problem, we use real-to-sim calibration, shown in Fig.~\ref{fig:method} (b), as an efficient source of behavior
videos.

Let \(\mathcal{D}_{\mathrm{R}}\) denote the real-world evaluation
distribution and \(\mathcal{D}_{\mathrm{S}}\) its calibrated simulation
counterpart. The goal is not to build a perfectly photorealistic digital twin,
but to calibrate the factors that determine policy behavior, including robots, tasks, cameras, and initializations. The simulated robot matches the real geometry, kinematics, joint limits, and
control interface. The simulated tasks match their real-world counterparts, and objects
are reconstructed as 3D assets and placed according to pose estimates. The camera viewpoints are calibrated to match the real geometry.

We then obtain evaluation videos through three stages.

\noindent\textbf{Adaptive policy training.}
Although calibration narrows the real-to-sim gap, residual differences in
friction, compliance, and sensing noise remain unavoidable
in contact-rich manipulation. Direct deployment in simulation may cause a policy to exhibit unstable behavior, making
the collected rollouts reflect residual domain mismatch rather than policy differences of interest. We therefore acquire calibrated simulation trajectories through action replay, motion planning, or
teleoperation, and adapt candidate policies. Let \(\pi_i^{\mathrm{R}}\) denote the policy evaluated in
the real world and \(\pi_i^{\mathrm{S}}\) its
simulation-adapted counterpart derived from the same underlying candidate.
R2S-Eval directly evaluates \(\pi_i^{\mathrm{S}}\) and does not assume that it
is identical to \(\pi_i^{\mathrm{R}}\). Instead, the comparative
ranking induced by \(\{\pi_i^{\mathrm{S}}\}\) is used as a proxy for the
hardware ranking of \(\{\pi_i^{\mathrm{R}}\}\). Sec.~\ref{real2sim_calibrate_exp} assesses this proxy
through closed-loop task performance and ranking consistency.

\noindent\textbf{Closed-loop simulation deployment.}
Each policy \(\pi_i^{\mathrm{S}}\) is deployed in closed loop in the calibrated
simulation. Evaluation videos are collected from the policy's own executions.

\noindent\textbf{Rollout video collection.}
For each \(\pi_i^{\mathrm{S}}\), we record
\begin{equation}
    \mathcal{V}_i=\{v_i^1,\ldots,v_i^{K_i}\},
\end{equation}
which serves as input to VLM preference evaluation. With calibrated simulation,
R2S-Eval reduces the need for repeated hardware trials.

\subsection{VLM Preference Evaluation}
\label{vlm_eval}

\begin{figure}[t]
    \centering
    \includegraphics[width=0.9\linewidth]{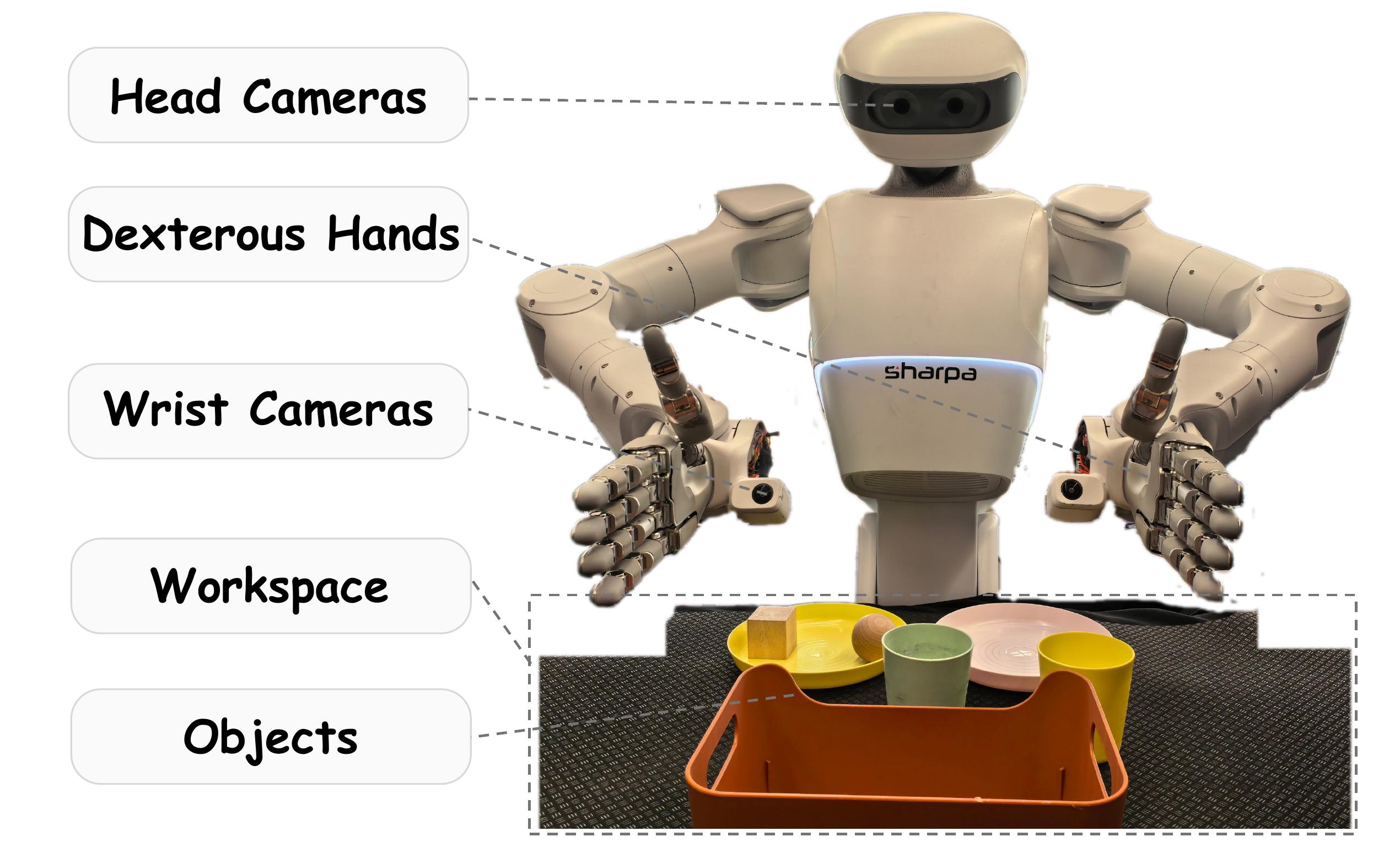}
    \vspace{-0.5em}
    \caption{SharpaNorth robot platform.}
    \label{fig:platform}
\vspace{-0.5em}
\end{figure}

Given the video sets \(\{\mathcal{V}_i\}_{i=1}^{N}\), VLM preference evaluation, shown in Fig.~\ref{fig:method} (c), estimates the preference observations. Humans watch videos and compare the overall quality
of execution, serving as a natural reference. However, this introduces substantial manual effort. We therefore use a
vision-language model (VLM) as an automated video-based preference judge. The
VLM is used not as a hand-designed task scorer but as a general visual judge in a way intended to match human preference. We evaluate this human agreement in
Sec.~\ref{sec:vlm_eval_results}. The module consists of 3 stages.

\noindent\textbf{Pairwise video sampling.}
For each pair \((\pi_i,\pi_j)\), we sample
\begin{equation}
    (v_i^a,v_j^b), \quad
    v_i^a \in \mathcal{V}_i,\; v_j^b \in \mathcal{V}_j .
\end{equation}
Each comparison uses videos under the
same task and initial configuration, and each task contributes the same number of comparisons. Repeated comparisons over randomly sampled task configurations therefore estimate preference between policy rollout distributions without confounding policy quality with task frequency or condition difficulty.

\noindent\textbf{Structured VLM judgment.}
Given a video \(v_i^a\) and its instruction \(x\), the VLM gives a structured behavior description
\begin{equation}
    d_i^a = G_{\mathrm{desc}}(v_i^a,x).
\end{equation}
The description covers motion smoothness, temporal continuity, task progress, and visually observable contact and effort control in the video. For a pair
\((v_i^a,v_j^b)\), the two anonymized videos are randomly assigned to positions
\(A\) or \(B\). The VLM compares the corresponding descriptions
under the same criteria and enforces a binary preference output
\begin{equation}
    y_{ij}^{ab}
    =
    G_{\mathrm{pref}}(d_i^a,d_j^b)
    \in \{i\succ j,\;j\succ i\}.
\end{equation}
This two-step process grounds the pairwise judgment in explicit execution evidence rather than a success signal. The task-level win counts are pooled into a preference matrix \(M\).

\noindent\textbf{Bradley--Terry preference aggregation.}
To obtain the policy-level ranking from \(M\), we instantiate the statistical preference model with the Bradley--Terry model~\cite{bradley1952rank}. The probability
that \(\pi_i\) is preferred over \(\pi_j\) is modeled as
\begin{equation}
    q_{ij}(\theta)
    =
    \Pr_{\theta}(\pi_i \succ \pi_j)
    =
    \frac{\exp(\theta_i)}
    {\exp(\theta_i)+\exp(\theta_j)}
    =
    \sigma(\theta_i-\theta_j).
\end{equation}
We fit BT strength parameters \(b_i>0\) using the
minorization-maximization algorithm~\cite{10.1214/aos/1079120141} and impose
\(\sum_i b_i=1\) for identifiability, with
\(\theta_i=\log b_i\). Using the likelihood objective, we estimate the policy scores \(\hat{\theta}\) and rank the policies accordingly.

\subsection{Validation Protocol}
\label{validation}

We complement the operational pipeline with a validation protocol in Fig.~\ref{fig:method} (d) along five axes.

\noindent\textbf{Reliability.}
A reliable evaluator should produce policy conclusions consistent with the target setting. For policy ranking, it should agree with a reference ranking that reflects the actual capabilities of the same set of policies.

\noindent\textbf{Stability.}
As evaluation evidence accumulates, the estimated policy scores or rankings should converge and exhibit low variance across different trials and configurations.

\noindent\textbf{Human agreement.}
The preference signal produced by the evaluator should agree with human judgments, indicating that its decisions reflect human assessment rather than an arbitrary model-specific scoring rule.

\noindent\textbf{Human effort.}
Manual work may be required to obtain an evaluation result, including hardware execution, scene reset, and monitoring. An accurate evaluator is still impractical if it requires substantial human intervention.

\noindent\textbf{Behavioral informativeness.}
Behavioral information should be preserved. Binary success records only completion and discards other details. An informative evaluator should capture differences in progress, recovery, or control quality, especially given the same success or failure outcome.

\section{Experiments}

In this section, we conduct simulation and real-world experiments to answer the following questions:
\emph{Q1.} Does the calibrated simulation preserve hardware-consistent policy conclusions?
\emph{Q2.} Does R2S-Eval produce policy conclusions that reflect the actual performance of candidate policies?
\emph{Q3.} Can R2S-Eval mitigate the instability and manual cost of conventional real-world evaluation?
\emph{Q4.} Do VLM judgments agree with human preferences?
\emph{Q5.} Does preference-based evaluation capture execution-level information?

\subsection{Real-to-Sim Calibration Results}
\label{real2sim_calibrate_exp}

\begin{table}[t]
\caption{Data statistics on real-world tasks.}
\label{tab:tasks}
\centering
\setlength{\tabcolsep}{4pt}
\begin{tabular*}{\columnwidth}{@{\extracolsep{\fill}}lccc@{}}
\toprule
Task & \# Real Traj. & \# Sim. Traj. & Replay Succ. (\%)\\
\midrule
Cover Ball with Cup       & 510 & 494 & 96.9 \\
Pick Ball Place in Box    & 497 & 491 & 98.8 \\
Pick Ball Place in Cup    & 509 & 491 & 96.5 \\
Pick Cup Place on Plate   & 502 & 502 & 100.0 \\
Stack Blocks              & 505 & 471 & 93.3 \\
Take Block out Box        & 503 & 499 & 99.2 \\
Upright Fallen Cup        & 499 & 497 & 99.6 \\
\midrule
\textbf{Total / Avg.}     & \textbf{3525} & \textbf{3445} & \textbf{97.8} \\
\bottomrule
\end{tabular*}
\vspace{0.0em}
\end{table}

\begin{table}[t]
\caption{Real--sim success rates on 7 tasks and 6 VLAs.}
\label{tab:real_sim_consistency}
\centering
\setlength{\tabcolsep}{4pt}

\begin{adjustbox}{max width=\columnwidth}
\begin{tabular}{@{}lcccccc@{}}
\toprule
Task & X-VLA & $\pi_{0.5}$ & $\pi_0$ & OpenVLA & Nora-Long & SmolVLA \\
\midrule
Cover Ball Cup  & 60.0 / 61.0 & 55.0 / 57.0 & 55.0 / 60.5 & 5.0 / 3.5 & 5.0 / 3.5 & 5.0 / 1.5 \\
Pick Ball Box   & 70.0 / 67.0 & 70.0 / 70.0 & 55.0 / 58.5 & 5.0 / 4.0 & 5.0 / 3.0 & 0.0 / 1.0 \\
Pick Ball Cup   & 50.0 / 48.0 & 50.0 / 52.0 & 45.0 / 52.0 & 5.0 / 3.0 & 5.0 / 2.5 & 0.0 / 0.5 \\
Pick Cup Plate  & 80.0 / 79.0 & 75.0 / 75.5 & 60.0 / 63.5 & 10.0 / 8.5 & 5.0 / 6.0 & 5.0 / 2.5 \\
Stack Blocks    & 45.0 / 46.5 & 45.0 / 45.5 & 40.0 / 42.5 & 5.0 / 2.5 & 0.0 / 1.5 & 0.0 / 0.5 \\
Take Block Box  & 65.0 / 67.5 & 65.0 / 65.0 & 55.0 / 58.0 & 5.0 / 4.5 & 5.0 / 2.5 & 0.0 / 1.0 \\
Upright Cup     & 90.0 / 85.5 & 85.0 / 86.5 & 65.0 / 71.0 & 15.0 / 11.0 & 5.0 / 7.5 & 5.0 / 4.0 \\
\midrule
\textbf{Total / Avg.}
& \textbf{65.7 / 64.9} & \textbf{63.6 / 64.5} & \textbf{53.6 / 58.1} & \textbf{7.1 / 5.3} & \textbf{4.3 / 3.7} & \textbf{2.1 / 1.6}\\
\bottomrule
\end{tabular}
\end{adjustbox}

\vspace{-0.5em}
\end{table}

This subsection answers \emph{Q1} by introducing the real-world experimental setup and assessing the real--sim consistency.

\noindent\textbf{Robot platform.}
All real-world experiments are conducted on SharpaNorth
, a dual-arm humanoid manipulation platform shown in Fig.~\ref{fig:platform}. The robot has two 7-DoF arms, each with a 22-DoF SharpaWave 
dexterous hand, as well as a 2-DoF neck, a 3-DoF upper body, and a 2-DoF waist. We keep the waist fixed, giving an active configuration of \(63\) DoFs. The observation system consists of two head-mounted RGB cameras for stereo workspace views and two wrist-mounted fish-eye cameras for close-range hand-object observations.

\noindent\textbf{Task suite.}
We design seven tabletop tasks shown in Fig.~\ref{fig:tasks}. The manipulated objects are randomly initialized on the tabletop within a predefined range. Task objects, initialization ranges, and success conditions remain consistent between the real world and the corresponding simulation.

\noindent\textbf{VLA policies.}
We evaluate six candidate VLA policies:
(1) \textbf{\(\boldsymbol{\pi_0}\)}~\cite{black2024pi0},
a flow-matching VLA built on a pretrained vision-language model;
(2) \textbf{\(\boldsymbol{\pi_{0.5}}\)}~\cite{intelligence2025pi05},
an open-world generalization variant of \(\pi_0\);
(3) \textbf{OpenVLA}~\cite{kim2024openvla},
a 7B open source VLA pretrained on large-scale robot demonstrations;
(4) \textbf{NORA-Long},
a long-action variant of
NORA~\cite{hung2025nora},
which adopts Qwen2.5-VL-3B~\cite{bai2025qwen25vl}
and the FAST+ action tokenizer~\cite{pertsch2025fast};
(5) \textbf{SmolVLA}~\cite{shukor2025smolvla},
a lightweight VLA for efficient training and deployment;
and
(6) \textbf{X-VLA}~\cite{zheng2025x},
a VLA with specific soft prompts for cross-embodiment
policy learning.

\noindent\textbf{Real-world data collection.}
We collect real-world trajectories with a teleoperation system, composed of an upper-body exoskeleton, exoskeleton gloves, and a VR headset. The exoskeleton tracks the operator's arm and hand motions and maps them to the robot action space, while the VR headset provides stereo visual feedback from the robot's head-mounted cameras. During teleoperation, we record synchronized head-camera views, wrist-camera views, robot proprioception, and executed actions. We collect 3,525 trajectories in total, which are shown in Tab.~\ref{tab:tasks}.

\begin{figure}[!t]
    \centering
    \includegraphics[width=1.0\linewidth]{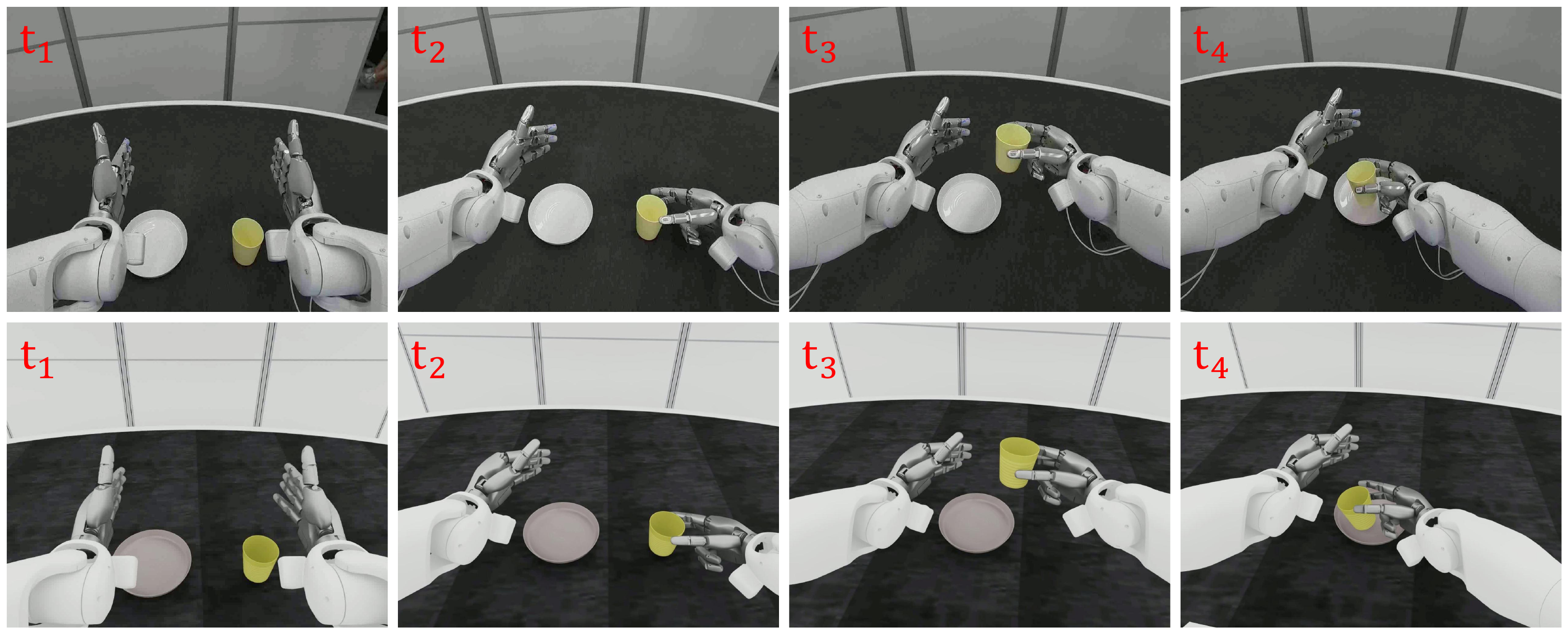}
    \vspace{-0.5em}
    \caption{A real-world trajectory (top) and its simulated action-replay counterpart
    (bottom) for the Pick Cup and Place on Plate task. Each row shows four key
    frames ordered from left to right.}
    \label{fig:real2sim}
\vspace{-0.5em}
\end{figure}

\noindent\textbf{Simulated data acquisition.}
Following Sec.~\ref{real2sim}, we build the simulation in NVIDIA Isaac Sim~\cite{nvidia2025isaacsim}. We
replay the recorded action sequences in the calibrated simulator to acquire the simulated data. Action replay is
sensitive to the initial poses of objects, as small errors can change contact timing and cause a real-world trajectory to fail in simulation. We
therefore use an automatic replay loop. If a replay attempt fails, we resample a perturbation around the estimated initial object pose and execute the same action sequence again until the replay succeeds or the retry limit is reached. Fig.~\ref{fig:real2sim} compares representative real
and simulated trajectories, and Tab.~\ref{tab:tasks} reports the trajectory counts and the success rates of action replay. 
The real-world policy \(\pi_i^{\mathrm{R}}\) and its simulation-side counterpart
\(\pi_i^{\mathrm{S}}\) are trained independently on real and simulated data, respectively, using identical
task definitions, observation and action interfaces, and the model-specific training recipe. 

\noindent\textbf{Closed-loop real--sim consistency.}
We compare the results of \(\pi_i^{\mathrm{R}}\) and \(\pi_i^{\mathrm{S}}\) from independently executed closed-loop policy rollouts. Tab.~\ref{tab:real_sim_consistency} reports each real/sim success rate. We run 20 and 200 trials per task in the real world and in simulation, respectively. The two routes produce closely matched
task-level success rates and the same overall policy ranking. Across all policy--task pairs, the mean absolute
real--sim difference is 2.13 percentage points. The largest gap is 7.0 percentage points, indicating that calibration does not eliminate all discrepancies.
In both domains, \(\pi_0\), \(\pi_{0.5}\), and X-VLA clearly outperform the remaining policies, which may be limited by model capacity or action representation. Real and simulated rollouts also exhibit qualitatively similar failure
modes, including near-static oscillations without progress and
large-amplitude oscillations during motion. These results do not imply behavioral equivalence between \(\pi_i^{\mathrm{R}}\) and
\(\pi_i^{\mathrm{S}}\), but show that the calibrated simulation is sufficient
to preserve comparative policy conclusions.

\subsection{VLM Preference Evaluation Results}
\label{sec:vlm_eval_results}

\begin{figure}[t]
    \centering
    \includegraphics[width=1.0\linewidth]{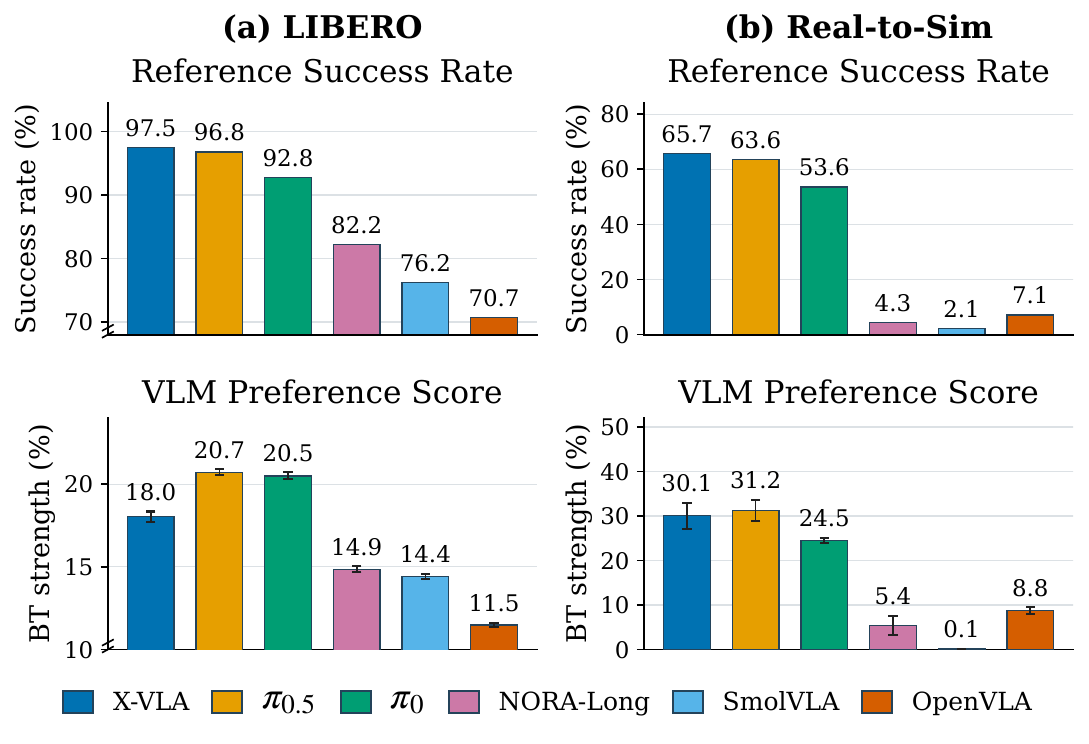}
    \vspace{-0.5em}
    \caption{Overall policy performance of LIBERO and Real-to-Sim experiments. The VLM preference score is averaged over all 8 VLMs.}
    \label{fig:ranking}
\vspace{-0.5em}
\end{figure}

This subsection addresses \emph{Q2}, the stability aspect of \emph{Q3}, and \emph{Q4} by evaluating the VLM preference evaluation results in simulated and calibrated real-to-sim settings.

\noindent\textbf{Evaluation metrics.}
For \emph{Q2}, we compare the results of the VLM with a reference performance. 
(1) \textbf{Spearman's \(\boldsymbol{\rho}\)} measures the agreement between the VLM ranking and the reference ranking. 
(2) \textbf{Pearson \(\boldsymbol{r}\)} measures the correlation between VLM policy scores and reference performance values. 
(3) \textbf{MMRV} (Mean Maximum Rank Violation)~\cite{li2024evaluating} measures the severity of ranking inversions.
For \emph{Q3}, we recompute policy scores and rankings as more video comparisons are incorporated. 
(4) \textbf{BT CI} measures the average width of the 95\% bootstrap confidence interval for standardized policy scores on a 0--100 scale.
For \emph{Q4}, we compare VLM preferences with human annotations. 
(5) \textbf{Human agreement rate} measures the percentage of pairs for which the VLM selects the same preferred video as human annotators. 

\noindent\textbf{Implementation details.}
(1) Reference performance is defined as the mean success rate on all tasks. (2) Bootstrap confidence intervals are estimated from 1,000 replicates. In each replicate, the number of comparisons is preserved, and the BT model is refitted. (3) Three human annotators view 200 video pairs and select the preferred ones under the same evaluation protocol as VLM. The sample pairs are approximately balanced across policies and outcome categories. The majority vote defines the human reference preference. (4) The metrics are averaged across checkpoints ranging from 500 to 2,000 video
comparisons per policy pair in increments of 100.

\noindent\textbf{VLM evaluators.}
We evaluate eight VLMs.
(1) \textbf{Qwen3-VL}~\cite{bai2025qwen3vltechnicalreport}
(4B, 8B, 32B),
(2) \textbf{LLaVA-OV-1.5}~\cite{an2025llavaonevision15}
(4B, 8B), and
(3) \textbf{Gemma 3}~\cite{gemmateam2025gemma3technicalreport}
(4B, 12B, 27B).

\noindent\textbf{Simulation results.}
The LIBERO~\cite{liu2023liberobenchmarkingknowledgetransfer} results are summarized in Fig.~\ref{fig:ranking} (a) and Tab.~\ref{tab:libero_vlm}. For each policy, we collect 2,000 rollout videos, 50 per task. 
Generally, the VLM ranking closely follows the success-rate ranking, with minor inversions among nearly tied policies. For example, \(\pi_{0.5}\) receives a higher score than X-VLA despite a slightly lower success rate, reflecting its better execution quality. Across all VLMs, the inferred policy rankings are highly consistent, achieving an average Spearman's \(\rho\) of 0.823, Pearson \(r\) of 0.924, and MMRV of 0.018. The narrow BT confidence intervals indicate robust ranking estimation with low uncertainty. The human agreement rate reaches 82.9\%, demonstrating that VLM judgments agree with human assessments. All VLMs exhibit similar performance despite differences in architecture and model scale, suggesting that the proposed evaluation pipeline generalizes well across diverse VLM backbones.

\begin{table}[t]
\caption{LIBERO VLM preference evaluation results.}
\label{tab:libero_vlm}
\centering
\footnotesize
\setlength{\tabcolsep}{3pt}
\begin{tabular*}{\columnwidth}{@{\extracolsep{\fill}}lccccc@{}}
\toprule
VLM & \(\rho\uparrow\)
& \(r\uparrow\)
& MMRV\(\downarrow\)
& BT CI\(\downarrow\)
& Agr. rate (\%)\(\uparrow\) \\
\midrule
Qwen3-VL-4B
& 0.813 & 0.915 & 0.019 & 1.231 & 80.5 \\
Qwen3-VL-8B
& 0.823 & 0.926 & 0.018 & 1.312 & 83.5 \\
Qwen3-VL-32B
& 0.813 & 0.935 & 0.020 & 1.429 & 87.0 \\
LLaVA-OV-1.5-4B
& 0.829 & 0.929 & 0.017 & 1.414 & 85.0 \\
LLaVA-OV-1.5-8B
& 0.829 & 0.908 & 0.017 & 1.346 & 83.0 \\
Gemma3-4B
& 0.829 & 0.903 & 0.017 & 1.285 & 78.0 \\
Gemma3-12B
& 0.829 & 0.941 & 0.017 & 1.392 & 84.0 \\
Gemma3-27B
& 0.823 & 0.939 & 0.018 & 1.285 & 82.0 \\
\midrule
\textbf{Avg.}
& \textbf{0.823} & \textbf{0.924} & \textbf{0.018} & \textbf{1.337} & \textbf{82.9} \\
\bottomrule
\end{tabular*}
\vspace{0.0em}
\end{table}

\begin{table}[t]
\caption{Real-to-sim VLM preference evaluation results.}
\label{tab:real2sim_vlm}
\centering
\setlength{\tabcolsep}{3pt}
\begin{tabular*}{\columnwidth}{@{\extracolsep{\fill}}lccccc@{}}
\toprule
VLM & \(\rho\uparrow\) & \(r\uparrow\) & MMRV\(\downarrow\) & BT CI\(\downarrow\) & Agr. rate (\%)\(\uparrow\) \\
\midrule
Qwen3-VL-4B        & 0.943 & 0.957 & 0.007 & 1.712 & 90.0 \\
Qwen3-VL-8B        & 0.943 & 0.984 & 0.007 & 1.750 & 91.5 \\
Qwen3-VL-32B       & 1.000 & 0.973 & 0.000 & 1.771 & 92.0 \\
LLaVA-OV-1.5-4B    & 0.943 & 0.989 & 0.006 & 1.827 & 88.5 \\
LLaVA-OV-1.5-8B    & 1.000 & 0.984 & 0.000 & 1.807 & 90.0 \\
Gemma3-4B          & 0.943 & 0.981 & 0.005 & 1.725 & 92.0 \\
Gemma3-12B         & 0.943 & 0.968 & 0.009 & 1.722 & 93.0 \\
Gemma3-27B         & 0.943 & 0.980 & 0.007 & 1.741 & 90.5 \\
\midrule
\textbf{Avg.} & \textbf{0.957} & \textbf{0.978} & \textbf{0.005} & \textbf{1.757} & \textbf{91.9} \\
\bottomrule
\end{tabular*}
\vspace{-0.5em}
\end{table}

\noindent\textbf{Real-to-sim results.}
The calibrated real-to-sim results are shown in Fig.~\ref{fig:ranking} (b) and Tab.~\ref{tab:real2sim_vlm}. All policies use identical camera, rendering, and temporal-sampling settings, reducing the risk that policy-specific visual artifacts serve as preference cues. For each policy, we collect 1,400 rollout videos, 200 per task. The reference performance is computed in the real world. As policies exhibit larger performance differences on the real robot, the inferred rankings are more consistent than those on LIBERO. Notably, \(\pi_{0.5}\) and X-VLA remain nearly tied across all evaluated VLMs, with some VLMs marginally favoring X-VLA and others slightly preferring \(\pi_{0.5}\). This behavior is consistent with their nearly identical real-world success rates, differing only by 2.1 percentage points. This indicates that the observed ranking differences reflect the small performance gap between the two policies.

\noindent\textbf{Order-swap consistency.}
For each real--sim video pair, the \emph{A/B} order is reversed and the evaluation is repeated. Across all VLMs, the preferred video remains unchanged for 91.5\% of the cases. We observe a slight positional bias when comparing nearly tied video pairs, where VLMs tend to favor the video presented as \emph{A}. To mitigate this, the two anonymized videos are randomly assigned to positions \emph{A} and \emph{B}, ensuring that any residual positional bias averages out over a large number of comparisons.

\subsection{Hardware-operation Effort}
\label{sec:manual_effort}

\begin{figure}[t]
    \centering
    \includegraphics[width=1.0\linewidth]{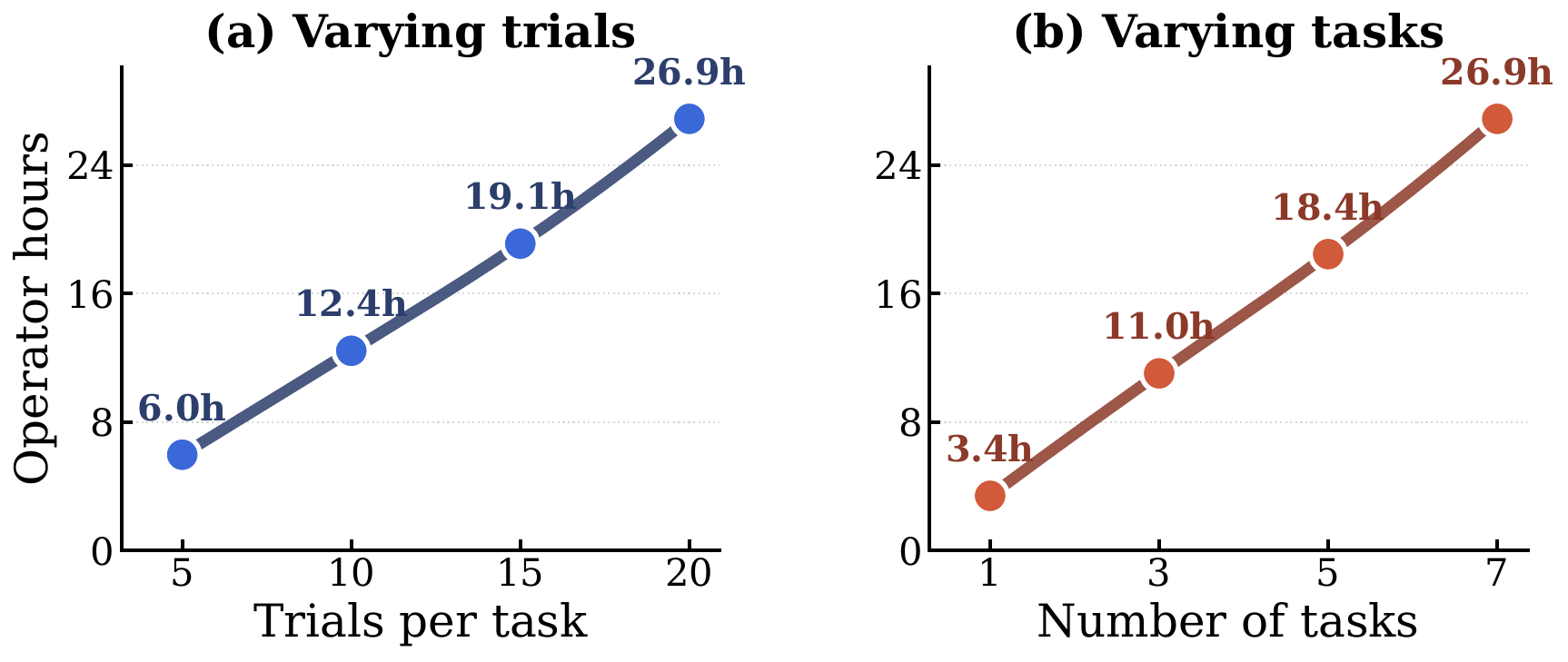}
    \vspace{-0.5em}
    \caption{Repeated hardware-operation effort.}
    \label{fig:manual_effort}
\vspace{-0.5em}
\end{figure}

This subsection addresses the manual effort aspect of \emph{Q3} by quantifying the repeated operator time required for the conventional real-world evaluation. We count hardware execution, object initialization and reset, monitoring, and success recording but exclude one-time data preparation and computational costs. Human annotation is used only for validation and is not required during routine evaluation.

We keep the six candidate policies fixed and sum the measured operator time, while varying either the number of task or trials per task. As shown in Fig.~\ref{fig:manual_effort}, the repeated hardware-operation time increases approximately linearly, while R2S-Eval does not require additional real-world hardware effort.

\subsection{Behavior Case Study}

To address \emph{Q5}, we examine rollout pairs with identical success labels. We select representative pairs in which both rollouts succeed or fail, and analyze whether VLM preference evaluation can distinguish their execution quality.

\begin{figure}[t]
    \centering
    \includegraphics[width=1.0\linewidth]{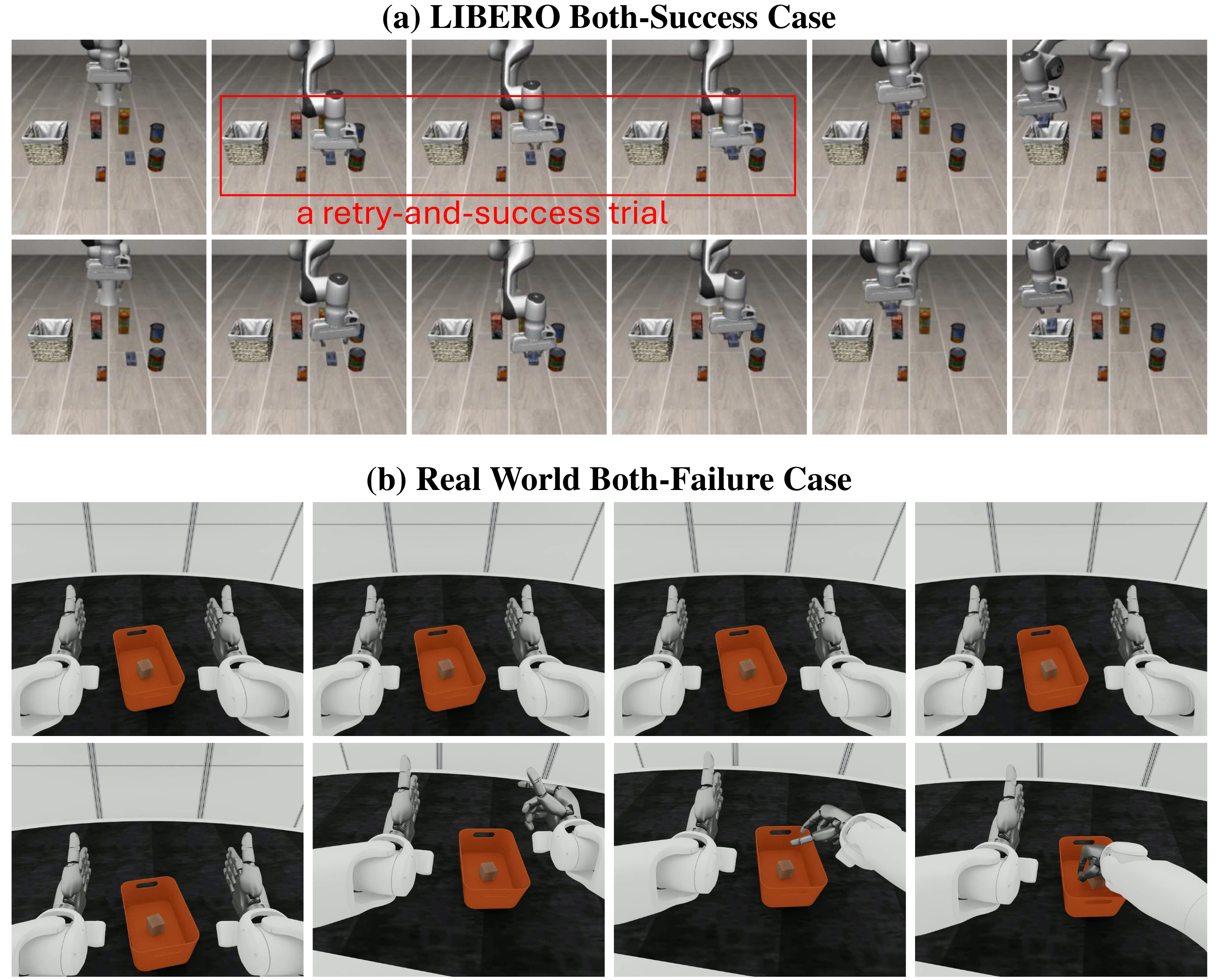}
    \vspace{-0.5em}
    \caption{Representative rollout pairs with identical binary outcomes.
    (a) Two successful rollouts for the LIBERO task ``Pick up the cream cheese and place it in the basket.'' 
    (b) Two failed rollouts for the real-to-sim task ``Take the Block out of the Box.'' 
    Each rollout proceeds from left to right.}
    \label{fig:case_study}
\vspace{-0.5em}
\end{figure}

In Fig.~\ref{fig:case_study} (a), both LIBERO rollouts are successful. The top grasps the object and places it smoothly in a single continuous execution, whereas the bottom fails its initial grasp and completes the task after a retry. In Fig.~\ref{fig:case_study} (b), both real-world rollouts fail to complete the task. The top reaches into the box and attempts to manipulate the object before failing, while the bottom exhibits only small oscillatory motions without any progress. In both cases, the VLM prefers the first rollout for its higher execution quality. These examples demonstrate that VLM preference evaluation uses cues from the full execution process, including task progress, motion continuity, and control quality, and can therefore capture behavior differences invisible to binary success labels.

\section{Conclusion}

This work introduced R2S-Eval, an evaluation pipeline for robot manipulation policies based on preferences over rollout behaviors. It combines real-to-sim calibration for efficient video collection with VLM preference evaluation for policy ranking. Simulation and real-world experiments show that R2S-Eval produces reliable and stable policy conclusions, agrees with human judgments, reduces manual effort, and captures execution-quality differences beyond success counts. We believe that R2S-Eval represents a step toward a scalable and behavior-aware evaluation of robot manipulation policies. Future work will extend R2S-Eval to broader tasks and robots, while systematically analyzing real-to-sim discrepancies and constructing higher-fidelity simulations.



\addtolength{\textheight}{-2\baselineskip}   
\balance
\bibliographystyle{IEEEtran}
\bibliography{reference}


\clearpage
\onecolumn
\fontsize{10pt}{12pt}\selectfont

\noindent{\LARGE\bfseries Appendix\par}
\vspace{1em}

\newcounter{appsec}
\newcounter{appsubsec}[appsec]

\newcommand{\appsection}[1]{%
  \refstepcounter{appsec}%
  \setcounter{appsubsec}{0}%
  \vspace{1.2em}%
  \noindent{\large\bfseries \Alph{appsec}. #1\par}%
  \vspace{0.6em}%
}

\newcommand{\appsubsection}[1]{%
  \refstepcounter{appsubsec}%
  \vspace{0.8em}%
  \noindent{\normalsize\bfseries
    \Alph{appsec}.\arabic{appsubsec}. #1\par}%
  \vspace{0.3em}%
}

\appsection{Real-World Setup}

\appsubsection{Hardware and Sensing}

All real-world experiments are conducted on SharpaNorth \footnote{\url{https://www.sharpa.com/pages/north}}
, a dual-arm humanoid manipulation platform. The robot has two 7-DoF arms, each equipped with a 22-DoF SharpaWave\footnote{\url{https://www.sharpa.com/pages/wave}}
dexterous hand, as well as a 2-DoF neck, a 3-DoF upper body, and a 2-DoF waist. The observation system consists of 2 head-mounted RGB cameras (1920×1536) and 2 wrist-mounted fish-eye cameras (480×384). The cameras operate at 30 Hz. The action stream stores the commanded target joint angles, while the state stream stores the measured joint angles. Both action and state are recorded at 60 Hz. Timestamps are provided for every stream to synchronize the 60 Hz proprioception with the 30 Hz observations.

\appsubsection{Action Space}
Tab.~\ref{tab:action_space} summarizes the unified action space. The unified action interface contains 65 joint targets. The two waist
dimensions are retained in the vector for interface compatibility, but are held
fixed during all experiments, leaving 63 actively controlled DoFs.
\begin{table}[!htbp]
\centering
\caption{Action space.}
\label{tab:action_space}
\begin{tabular}{@{}cllc@{}}
\toprule
Index & Component & Joints & Dim. \\
\midrule
$0$--$6$   & Left arm  & \texttt{joint\_1} \dots \texttt{joint\_7} & 7 \\
$7$--$13$  & Right arm & \texttt{joint\_1} \dots \texttt{joint\_7} & 7 \\
\midrule
\multicolumn{4}{@{}l}{\emph{Left hand (22-DoF), indices $14$--$35$}} \\
$14$--$18$ & \quad Thumb  & CMC\_FE, CMC\_AA, MCP\_FE, MCP\_AA, IP & 5 \\
$19$--$22$ & \quad Index  & MCP\_FE, MCP\_AA, PIP, DIP             & 4 \\
$23$--$26$ & \quad Middle & MCP\_FE, MCP\_AA, PIP, DIP             & 4 \\
$27$--$30$ & \quad Ring   & MCP\_FE, MCP\_AA, PIP, DIP             & 4 \\
$31$--$35$ & \quad Pinky  & CMC, MCP\_FE, MCP\_AA, PIP, DIP        & 5 \\
\midrule
\multicolumn{4}{@{}l}{\emph{Right hand (22-DoF), indices $36$--$57$}} \\
$36$--$40$ & \quad Thumb  & CMC\_FE, CMC\_AA, MCP\_FE, MCP\_AA, IP & 5 \\
$41$--$44$ & \quad Index  & MCP\_FE, MCP\_AA, PIP, DIP             & 4 \\
$45$--$48$ & \quad Middle & MCP\_FE, MCP\_AA, PIP, DIP             & 4 \\
$49$--$52$ & \quad Ring   & MCP\_FE, MCP\_AA, PIP, DIP             & 4 \\
$53$--$57$ & \quad Pinky  & CMC, MCP\_FE, MCP\_AA, PIP, DIP        & 5 \\
\midrule
$58$--$64$ & Torso / neck  & Upper body (3) + waist (2) + neck (2) & 7 \\
\midrule
\multicolumn{3}{@{}l}{\textbf{Total}} & \textbf{65} \\
\bottomrule
\end{tabular}
\end{table}

\appsubsection{Unified Policy Execution}

To ensure a fair comparison, all evaluated policies are provided with the same camera streams and are
executed through the same control interface, with identical control frequency,
initialization, and termination criteria. Each model consumes the subset of
available observations supported by its architecture, and the model-specific output is converted to the unified action representation before execution.

\appsubsection{Objects and Assets}

Real-world manipulation tasks involve a set of objects. To construct the calibrated simulation environment, simulation assets are created for all objects used in the experiments. For most manipulation objects (e.g., cups and plates), we reconstruct textured USD assets from the corresponding real objects through 3D scanning and texture mapping, where the object geometry is represented as a USD mesh and the appearance is preserved by texture maps. Simple scene elements and geometric objects, such as tables and walls, are manually modeled using primitive geometries.

\appsubsection{Tasks}

\begin{figure}[!htbp]
    \centering
    \includegraphics[width=1.0\textwidth]{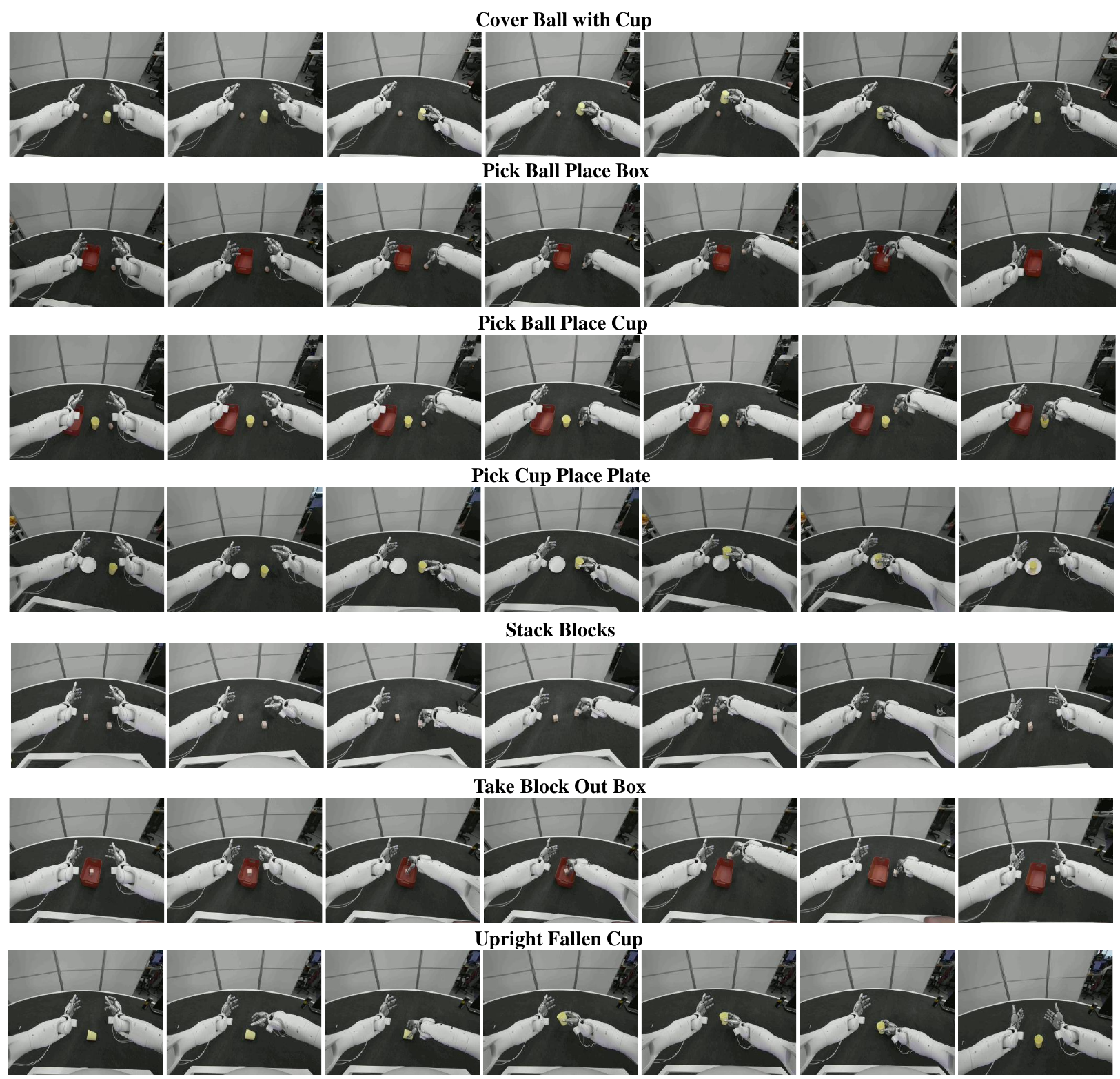}
    \caption{Seven real-world manipulation tasks. Each row shows seven frames from left to right.}
    \label{fig:more_tasks}
\vspace{-1.0em}
\end{figure}

The seven real-world tasks and their success criteria are defined below. Unless otherwise specified, all movable objects are randomly initialized by sampling their positions (x,y) and yaw orientations within predefined regions on the tabletop while avoiding initial collisions. All closed-loop evaluation episodes use newly sampled initializations and are disjoint from the training trajectories.
Task examples are shown in Fig.~\ref{fig:more_tasks}.

\noindent{\textbf{Cover the Ball with the Cup.}}
The robot grasps the cup and places it upside down to cover the ball. The cup is initialized upside down. The task is considered successful if the ball is completely covered by the cup at the end of the episode.

\noindent{\textbf{Pick the Ball and Place it in the Box.}}
The robot grasps the ball and places it inside the box. The task is considered successful if the ball is fully contained within the box.

\noindent{\textbf{Pick the Ball and Place it in the Cup.}}
The robot grasps the ball and places it inside the cup. The task is considered successful if the ball is fully contained within the cup.

\noindent{\textbf{Pick the Cup and Place it on the Plate.}}
The robot grasps the cup and places it upright on the plate. The task is considered successful if the cup is placed stably on the plate.

\noindent{\textbf{Stack the Two Blocks.}}
The robot grasps one block and stacks it on top of the other block. The task is considered successful if one block is stacked stably on top of the other.

\noindent{\textbf{Take the Block out of the Box.}}
The robot grasps the block from inside the box and places it outside the box. The task is considered successful if the block is completely removed from the box.

\noindent{\textbf{Upright the Fallen Cup.}}
The robot grasps the fallen cup and restores it to an upright pose. The cup is initially placed in a fallen configuration. The task is considered successful if the cup remains upright at the end of the episode.

\appsection{Real-to-Sim Calibration}
\appsubsection{Simulation Framework}

Real-to-sim calibration is implemented using the NVIDIA Isaac Lab-Arena framework \url{https://github.com/isaac-sim/IsaacLab-Arena}, an open-source extension to NVIDIA Isaac Lab for simplified task curation and robotic policy evaluation at scale.

\appsubsection{Simulation Scene Construction}

The simulation environment shares a common scene configuration for all tasks. The robot is instantiated from a calibrated USD articulation with the same kinematic structure as the real platform. Arms, lower body, and neck are controlled using implicit joint actuators, whereas dexterous hands use ideal PD actuators. Four RGB cameras are attached to the robot, including two head-mounted pinhole cameras ($1920\times1536$) and two wrist-mounted fisheye cameras ($480\times384$), which operate at approximately 30 Hz. Their mounting poses and projection models are configured according to the corresponding links and camera specifications of the physical robot.

The policy observation interface contains the four RGB streams, previous action, joint positions, and joint velocities. The images are rendered at their native camera resolutions and resized to \(224\times224\) during model preprocessing. At the beginning of each episode, all joints are reset to their predefined initial configuration. The articulation enables collision handling and self-collision, and uses actuator gains and dynamics parameters stored in the calibrated robot USD asset.

The surrounding scene consists of a custom tabletop, a background wall, and dome lighting. The table and wall are loaded as rigid kinematic bodies and placed to reproduce the layout of the real-world workspace.

Task-specific manipulation objects are instantiated from their corresponding simulation assets and assigned their initial poses before setting up the environment. All tasks therefore share the same robot, workspace, cameras, lighting, and rendering configuration, differing only in the manipulation objects and their initial states. The tabletop height can be adjusted for individual task configurations when necessary.

The simulator exposes the same observation modalities and control interface
as the real-world system for data collection and closed-loop evaluation.

Tab.~\ref{tab:sim_setup} summarizes the simulation parameters.

\begin{table}[!htbp]
\centering
\caption{Simulation environment configuration.}
\label{tab:sim_setup}
\begin{tabular}{@{}ll@{}}
\toprule
Item & Value \\
\midrule
Physics timestep       & $1/200$\,s (200\,Hz) \\
Render timestep        & $1/100$\,s (100\,Hz) \\
PhysX solver           & TGS (\texttt{solver\_type}=1) \\
Substeps               & None (position-iteration based) \\
Articulation pos./vel. iters & 8 / 0 \\
Bounce threshold vel.  & 0.5\,m/s \\
Friction offset thresh.\ / corr.\ dist. & 0.04 / 0.025 \\
Gravity                & $(0,0,-9.81)$\,m/s$^2$ \\
Object static / dynamic friction & 0.5 / 0.5 (Isaac Lab default) \\
Object restitution     & 0.0 (Isaac Lab default) \\
Robot contact offset / rest offset & 0.002 / 0.0\,m \\
Object contact / rest offset & PhysX default (not overridden) \\
Max depenetration vel.\ (robot) & 1000\,m/s \\
Self-collision (robot) & Enabled \\
\bottomrule
\end{tabular}
\end{table}

\appsubsection{Task Implementation}

\begin{figure}[!htbp]
    \centering
    \includegraphics[width=1.0\textwidth]{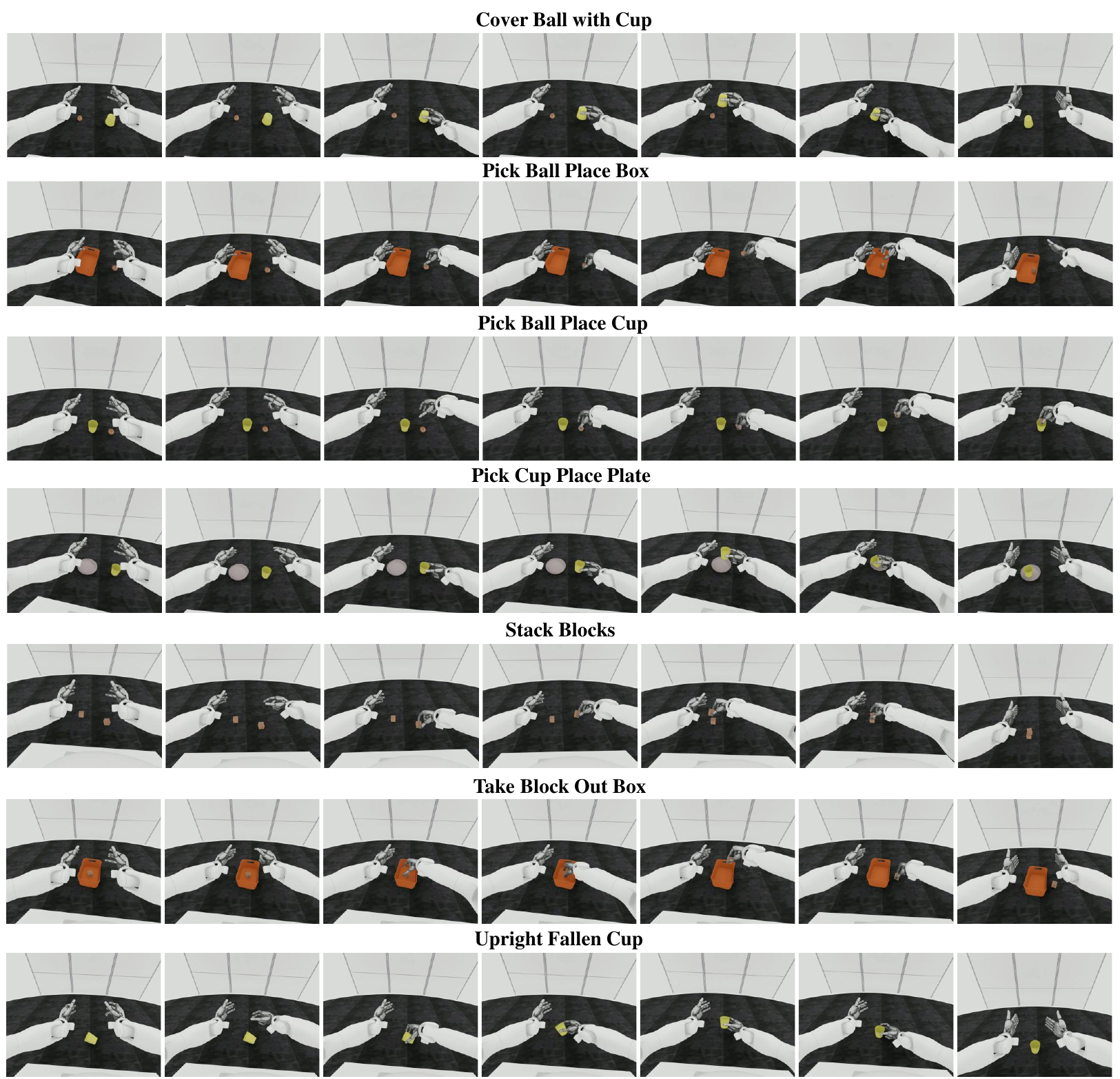}
    \caption{Seven simulated manipulation tasks. Each row shows seven frames from left to right.}
    \label{fig:sim_task_trajs}
\vspace{-1.0em}
\end{figure}

Each simulation task is implemented to match the task specification of its real-world counterpart. Each simulated task uses the same language instruction, initialization protocol, and episode configuration as its real-world counterpart. Unless otherwise specified, movable objects are initialized by randomly sampling their positions and yaw orientations within predefined regions while avoiding initial collisions. Task completion is determined using rule-based success evaluators that are consistent with the real-world evaluation protocol. Examples of simulated tasks are shown in Fig.~\ref{fig:sim_task_trajs}.

\appsubsection{Initialization Estimation}

We estimate the initial object poses in simulation through a three-stage procedure.

\noindent{\textbf{Segmentation of Real-World Objects.}} We perform open-vocabulary object segmentation using Grounding DINO for text-guided object detection and the Segment Anything Model (SAM) for mask refinement. The first RGB frame of each real-world trajectory is segmented to obtain object masks. Task-specific text prompts are used to detect all manipulation objects, from which the object centroids and principal orientations are extracted. These quantities provide the image-space observations required for initialization.

\noindent{\textbf{Real--Sim Pixel Alignment.}} The joint positions of the robot recorded at the first timestep are used to initialize the robot configuration in simulation. The object centroids are back-projected onto the tabletop using the calibrated camera intrinsics, extrinsics, and known tabletop height to recover their planar positions. The object yaw is estimated from the principal axis of each segmentation mask and transformed into the simulator world frame according to the camera pose. The initialized scene is then iteratively refined by minimizing the pixel reprojection error between the projected object centroids in the simulation and the corresponding centroids in the real image. At each iteration, the pixel reprojection error is converted into a correction in the tabletop coordinate frame through inverse projection, and the object positions are updated accordingly until convergence.

\noindent{\textbf{Task-Specific Error Calibration.}} For each task, a small set of trajectories, which are disjoint from all
closed-loop evaluation rollouts, is replayed in simulation to calibrate task-specific initialization offsets. The replayed trajectories are compared with their real-world counterparts, and small systematic discrepancies caused by object geometry, perception ambiguity, or simulator dynamics are compensated by manually adjusting the initialization parameters. The resulting calibration offsets are fixed for the task and subsequently applied to all policy trajectories of the same task without further per-trajectory tuning. 

\appsubsection{Action Replay}
We generate simulated trajectories by replaying the recorded joint-angle action sequences in the calibrated simulator. Replay outcomes are sensitive to small initialization errors because slight pose deviations may alter contact timing and cause a trajectory that succeeds in the real world to fail in simulation. We therefore employ an automatic replay procedure that searches within a small neighborhood of the estimated object poses. From the calibrated initialization, the recorded action sequence is executed and evaluated using the task-specific rule-based success criterion. If the replay fails, small pose perturbations are sampled around the estimated initialization and the trajectory is executed again. The procedure ends when a successful replay is obtained or the predefined retry limit is reached. Alg.~\ref{alg:action_replay} summarizes the automatic replay procedure. Fig.~\ref{fig:action_replay} presents additional examples of real-to-sim trajectory correspondence. We use \(N=8\) attempts. Translation offsets are sampled uniformly within
\([-0.01,0.01]\mathrm{m}\times[-0.01,0.01]\mathrm{m}\), and yaw offsets within
\([-\pi,\pi]\mathrm{rad}\).

\begin{algorithm}[!htbp]
\caption{Automatic Action Replay}
\label{alg:action_replay}
\begin{algorithmic}[1]
\Require Recorded action sequence $\mathbf{A}=\{\mathbf{a}_t\}_{t=1}^{T}$,
estimated initial object poses $\hat{\mathbf{P}}$,
retry limit $N$, perturbation distribution $\mathcal{D}$
\Ensure Replayed simulation trajectory $\tau_{\mathrm{sim}}$ and replay status

\For{$i = 1$ to $N$}
    \If{$i = 1$}
        \State $\mathbf{P}^{(i)} \gets \hat{\mathbf{P}}$
    \Else
        \State $\boldsymbol{\epsilon}^{(i)} \sim \mathcal{D}$
        \State $\mathbf{P}^{(i)} \gets \hat{\mathbf{P}} + \boldsymbol{\epsilon}^{(i)}$
    \EndIf
    \State Reset the simulator using $\mathbf{P}^{(i)}$
    \State Initialize the robot from the first recorded joint state
    \State $\tau_{\mathrm{sim}} \gets \Call{Execute}{\mathbf{A}}$
    \If{$\Call{SuccessEvaluator}{\tau_{\mathrm{sim}}}$}
        \State \Return $\tau_{\mathrm{sim}}, \mathrm{success}$
    \EndIf
\EndFor
\State \Return $\varnothing, \mathrm{failure}$
\end{algorithmic}
\end{algorithm}

\appsection{VLA Training and Deployment}

\appsubsection{Implementation Details}

\noindent{\textbf{LIBERO.}} Official publicly released checkpoints are used. No additional fine-tuning is needed. 

\noindent{\textbf{Real World.}} For real-world and real-to-sim experiments, all VLA policies are fine-tuned using the official open-source implementations and their recommended training configurations. For each VLA, a single multi-task policy is trained using demonstrations from all tasks, rather than training separate policies for individual tasks. None of the pretrained VLA policies natively supports the target action dimensionality. Therefore, the action projection layer is initialized by partially loading the pretrained weights, while the newly introduced dimensions are randomly initialized. For each VLA family, the real-world and simulation-side counterparts are
initialized from the same pretrained checkpoint and use identical
model-specific training recipes, update budgets, task definitions, and
observation/action interfaces. Training recipes may differ across VLA
families, following their respective official implementations. Unless otherwise specified, no architecture-specific modifications are introduced beyond task-dependent training hyperparameters. Specifically, the original OpenVLA autoregressively predicts the discrete token sequence of a single action at each policy query and does not natively support action chunking. To ensure a consistent action interface across all evaluated VLAs, we extend its autoregressive decoder to predict action chunks. The official repositories used in this work are listed below.

\begin{itemize}
    \item $\pi_{0}$: \url{https://github.com/Physical-Intelligence/openpi}
    \item $\pi_{0.5}$: \url{https://github.com/Physical-Intelligence/openpi}
    \item OpenVLA: \url{https://github.com/openvla/openvla}
    \item Nora-Long: \url{https://github.com/declare-lab/nora}
    \item X-VLA: \url{https://github.com/2toinf/X-VLA}
    \item SmolVLA: \url{https://github.com/huggingface/lerobot}
\end{itemize}

\appsubsection{Detailed Training Parameters}

The detailed training hyperparameters are shown in Tab.~\ref{tab:pi0_hyperparameters} for $\pi_{0}$, Tab.~\ref{tab:pi05_hyperparameters} for $\pi_{0.5}$, Tab.~\ref{tab:openvla_hyperparameters} for OpenVLA, Tab.~\ref{tab:nora_hyperparameters} for Nora-Long, Tab.~\ref{tab:xvla_hyperparameters} for X-VLA, and Tab.~\ref{tab:smolvla_hyperparameters} for SmolVLA. 

\appsubsection{Detailed Deployment Parameters}

In each policy query, every VLA predicts a 25-step action chunk. The first
five actions are executed before the policy is queried again. This receding-horizon execution protocol is used for all policies. 

\appsubsection{Computational Resources}

All VLA policies are fine-tuned on NVIDIA H20 GPUs until the training
objective stabilizes and no further improvement is
observed. Policy inference and simulation are conducted on NVIDIA
GeForce RTX 4090 GPUs.

\appsection{VLM Preference Evaluation}

\appsubsection{Prompt Templates}

The following prompt template is used for all VLM descriptions. The placeholder \texttt{\{video\_name\}} denotes the manipulation task corresponding to the input video. The actual input consists of the video together with its task name, but contains no policy, model, checkpoint, or outcome information.

\begin{lstlisting}[style=prompt]
You are a professional video analysis and robot behavior evaluation expert.

Your job is to describe the video {video_name} with a strict focus on robot motion quality.
Do NOT give a generic description. Do NOT summarize only the good parts. If there are flaws, even small ones, you must describe them explicitly.

You MUST describe the video through the following four dimensions:

1. Motion Smoothness
- Describe whether movements are steady, fluid, and continuous.
- Mention visible hesitation, jerky motions, jitter, pauses, or unnecessary adjustments.

2. Action Continuity / Temporal Coherence
- Describe whether the action sequence flows naturally from step to step.
- Mention if transitions are abrupt, repetitive, or uncoordinated.
- Identify any breaks in continuity or wasted motions.

3. Task Success / Goal Completion:
- Evaluate task success STRICTLY based on robot actions that are clearly initiated and actually completed in the video.
- Focus ONLY on what the robot demonstrably does, not on what could or should have been done.
- Describe concrete executed substeps (e.g., grasp, lift, place) and whether each substep is successfully completed.
- If any initiated substep is not completed (e.g., grasp attempted but not lifted or placed), you MUST treat it as a failure for that subtask.
- Explicitly state the final outcome using ONE of the following labels ONLY: successful, partially successful, or failed.
- Do NOT override failed or incomplete execution with an overall success judgment.
- Do NOT infer or assume any additional goals beyond the actions visibly attempted by the robot.
- Ignore untouched objects and background elements unless the robot clearly attempts and fails to interact with them.

4. Force / Effort Control:
- Describe how the robot applies force during contact, grasping, insertion, or placement.
- Indicate whether the applied force appears appropriate, excessive, insufficient, or inconsistent.
- Mention visible signs such as object wobbling, slipping, squeezing, pressing, or repeated force adjustments.
- Focus ONLY on observable effects of force application. Do NOT infer numerical force values.
- If force misapplication contributes to failure or hesitation, describe it explicitly.

Additional Rules:
- Use only what is visible in the video. Do not hallucinate.
- Always describe how the robot executes each key action.
- Use fine-grained temporal terms such as 'short pause', 'minor correction', 'smooth lift', 'slow transition', 'repeated adjustment', etc.
- Focus on the behavior of the robot, not the background unless relevant to the task.

Final Output Format:
Section 1: Motion Smoothness
Section 2: Action Continuity
Section 3: Task Success
Section 4: Force / Effort Control
Section 5: Short Overall Summary
\end{lstlisting}

The following prompt template is used for all VLM preference comparisons. The placeholder \texttt{\{desc\_block\}} denotes the
descriptions corresponding to the paired input videos.

\begin{lstlisting}[style=prompt]
You are an expert in robotic performance comparison. You must compare Robot A and Robot B based on action details, continuity, fluidity, and task success, and clearly determine which robot performs better.

{desc_block}
Your goal is to decide which video is better based ONLY on these criteria:
1. Motion Smoothness
2. Continuity / Fluidity
3. Task Completion
4. Force / Effort Control

Rules:
- Use ONLY the information explicitly stated in the descriptions.
- Do NOT hallucinate or infer unstated details.
- Smooth, continuous, steady actions → positive evidence.
- Jerky, hesitant, or multiple unnecessary adjustments → negative evidence.

Evaluation Procedure (you MUST follow internally, but DO NOT output any analysis):
1. Extract evidence for each video internally.
2. Score each dimension internally.
3. Internally decide which robot performs better.

Final Output Rule:
- Output ONLY one uppercase letter: "A" or "B".
- Do NOT output explanations, reasoning, analysis, or any other text.
\end{lstlisting}

\appsubsection{Detailed VLM Parameters}
The detailed VLM parameters for the generation of video descriptions and preference evaluation are summarized in Tab.~\ref{tab:vlm_inference}.
\begin{table}[!htbp]
\centering
\caption{VLM inference configurations.}
\label{tab:vlm_inference}
\begin{tabular}{ll}
\toprule
Parameter & Value \\
\midrule
Sampling FPS & 10 \\
Temperature & 0 \\
Max output tokens & 1024 \\
Video input & Full video \\
Generation strategy & Greedy ($\texttt{do\_sample=False}$) \\
Overlength handling   &   Uniform temporal subsampling / truncation \\
Policy identity    &      Hidden \\
A/B assignment     &      Randomized and balanced per policy pair\\
Tie handling       &      Not allowed; forced binary A/B output\\
Invalid outputs    &      Re-queried once / excluded and resampled\\
\bottomrule
\end{tabular}
\end{table}

\begin{figure}[!htbp]
\centering
\includegraphics[width=\columnwidth]{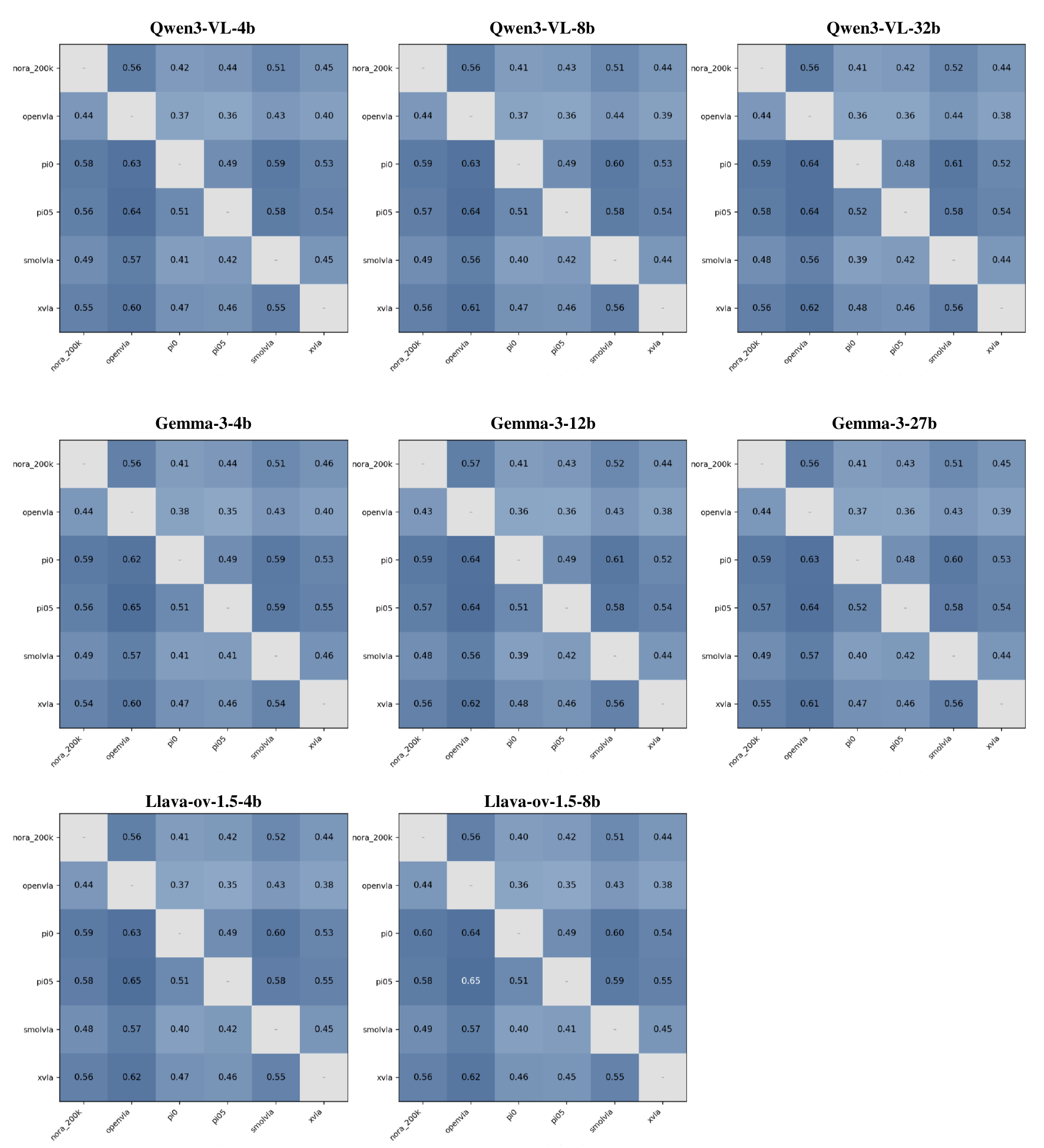}
\caption{LIBERO win-rate matrices obtained with different VLMs at the
2,000-comparison checkpoint.}
\label{fig:libero_preference_matrix}
\end{figure}

\begin{figure}[!htbp]
\centering
\includegraphics[width=\columnwidth]{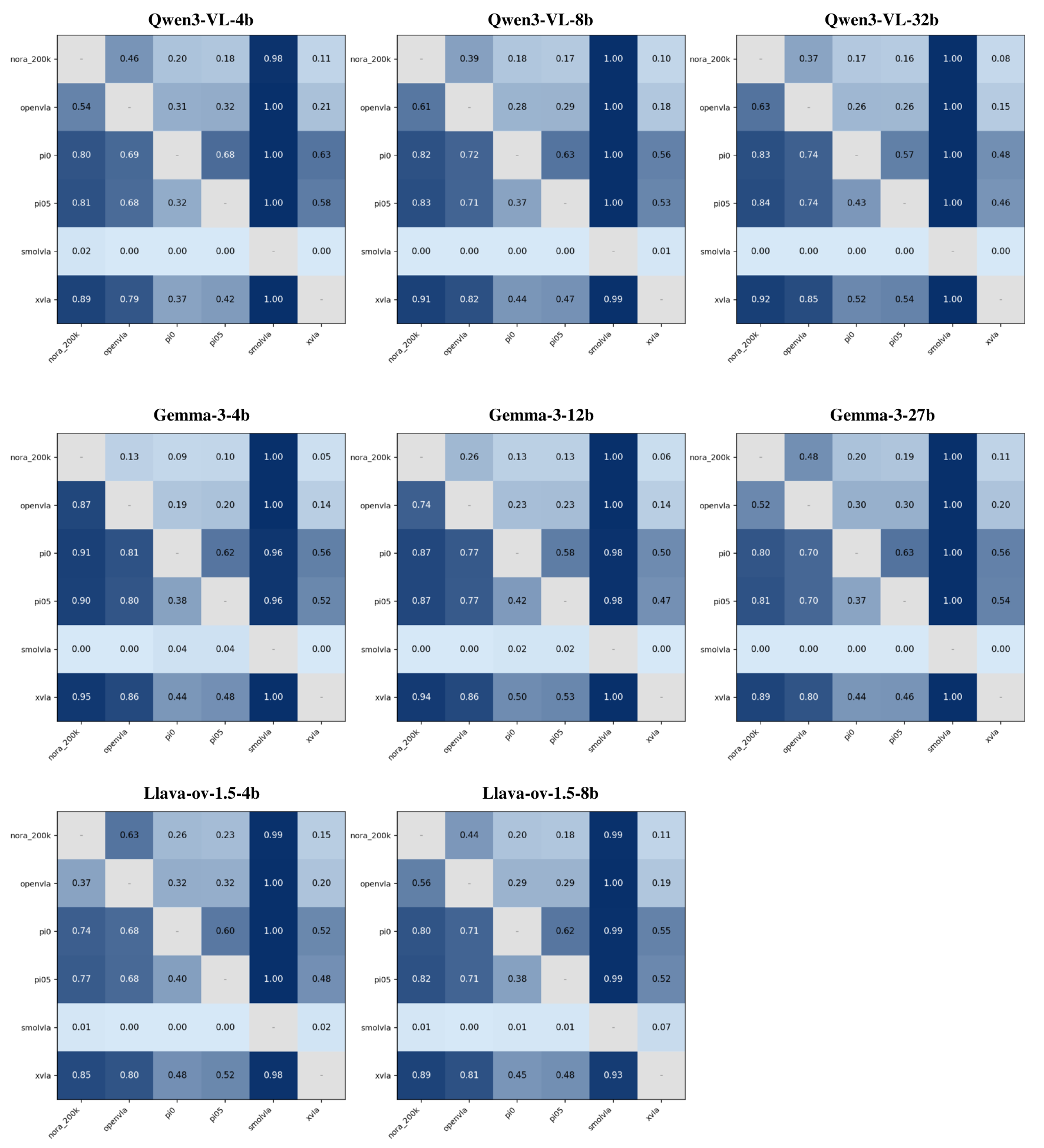}
\caption{Real-to-Sim win-rate matrices obtained with different VLMs at the
2,000-comparison checkpoint.}
\label{fig:realsim_preference_matrix}
\end{figure}

\appsubsection{Additional Results on Preference Matrix}
Fig.~\ref{fig:libero_preference_matrix} and Fig.~\ref{fig:realsim_preference_matrix} show the win-rate matrices obtained with different VLMs at the final 2,000-comparison checkpoint for LIBERO and real-to-sim settings, respectively.

\appsubsection{Detailed Bradley--Terry Model Parameters}

The detailed parameters of the Bradley--Terry model for preference aggregation are shown in Tab.~\ref{tab:bt_config}.

\begin{table}[!htbp]
\centering
\caption{Bradley--Terry model fitting and bootstrap configurations.}
\label{tab:bt_config}
\begin{tabular}{lc}
\toprule
Parameter & Value \\
\midrule
Optimization algorithm & MM \\
Maximum iterations & 1000 \\
Convergence tolerance & $10^{-8}$ \\
Numerical stability constant & $10^{-12}$ \\
Strength constraint    &   \(\sum_i b_i=1\)\\
Latent score        &      \(\theta_i=\log b_i\) \\
Reported score & \(s_i = 100 b_i\)\\
Bootstrap replicates & 1000 \\
Confidence level & 95\% \\
Bootstrap unit    &    Binomial resampling within each unordered policy pair \\
Comparison count   &   Preserved for every policy pair \\
Confidence interval &  95\% percentile interval on \(s_i\) \\
\bottomrule
\end{tabular}
\end{table}

\begin{figure}[!htbp]
\centering
\includegraphics[width=\columnwidth]{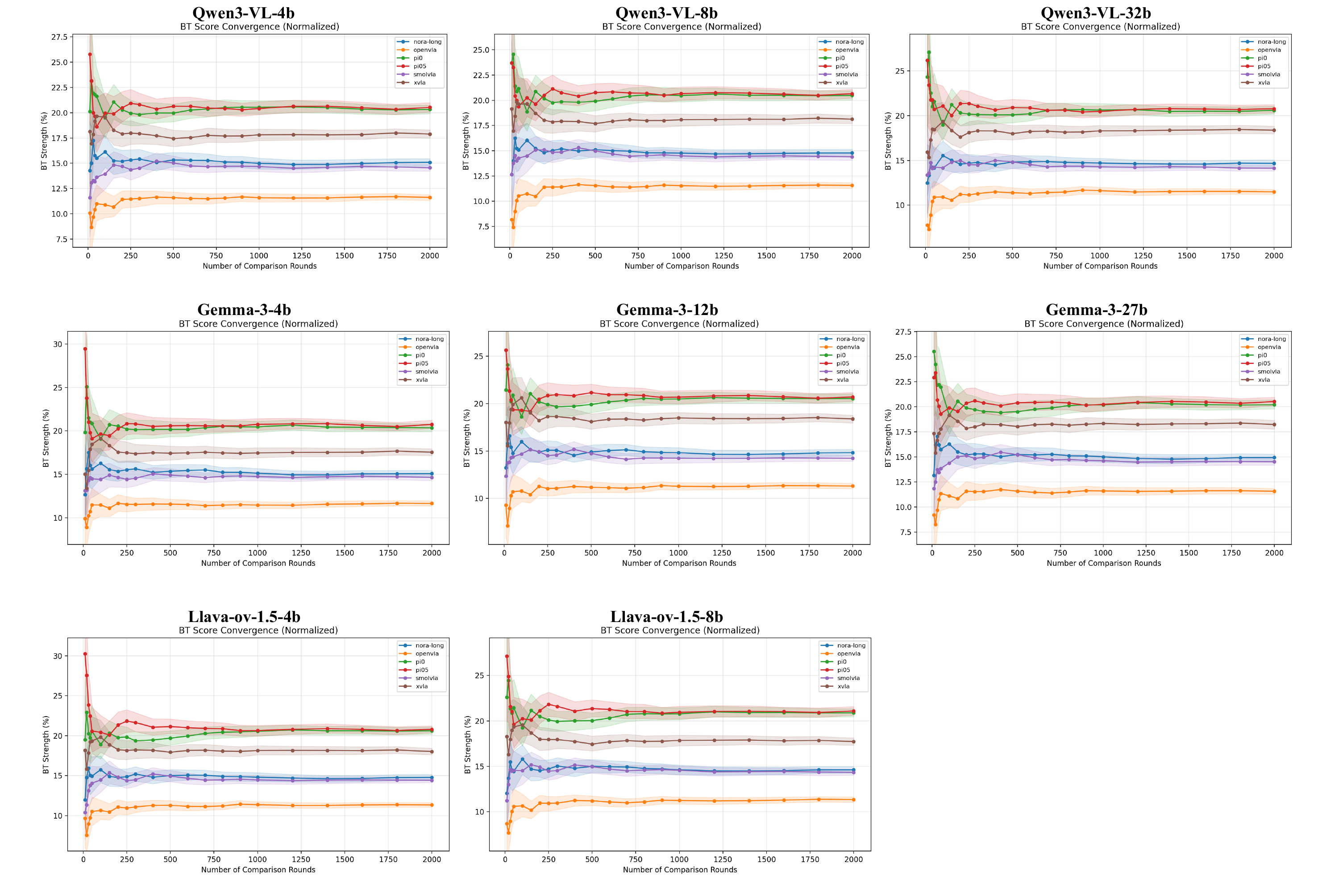}
\caption{Convergence curves of normalized Bradley--Terry scores for the six VLA policies under different VLM evaluators on LIBERO.}
\label{fig:libero_bt_convergence}
\end{figure}

\begin{figure}[!htbp]
\centering
\includegraphics[width=\columnwidth]{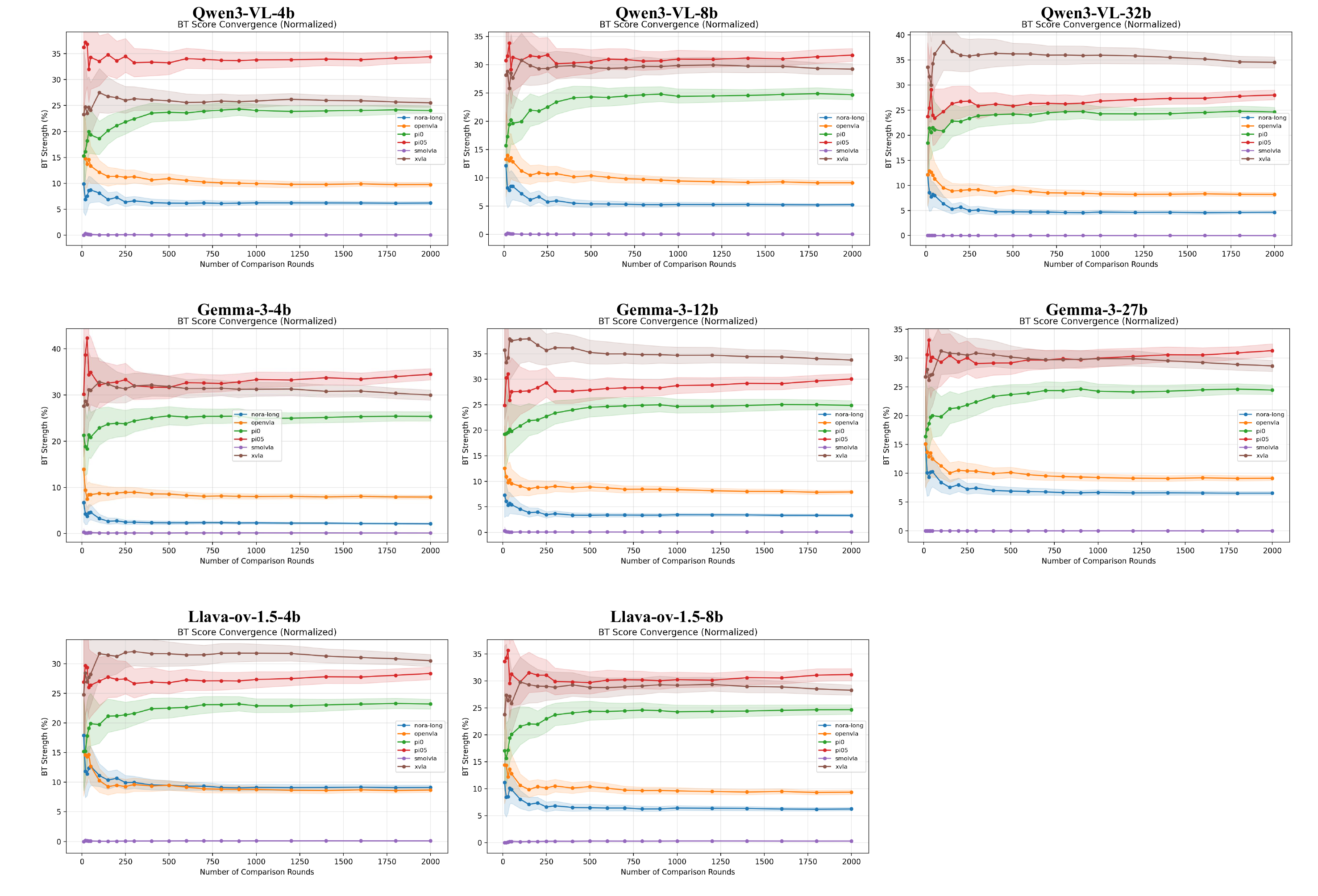}
\caption{Convergence curves of normalized Bradley--Terry scores for the six VLA policies under different VLM evaluators in the real-to-sim setting.}
\label{fig:realsim_bt_convergence}
\end{figure}

\appsubsection{Additional Results on Bradley--Terry Convergence Curves}
Detailed convergence curves on normalized Bradley–Terry model scores for all VLAs are shown in Fig.~\ref{fig:libero_bt_convergence} for LIBERO and Fig.~\ref{fig:realsim_bt_convergence} for real-to-sim.

\appsubsection{Additional Results on Bradley--Terry Scores}
Detailed normalized Bradley–Terry model scores for all VLAs are shown in Tab.~\ref{tab:bt_scores_libero} for LIBERO and in Tab.~\ref{tab:bt_scores_real_to_sim} for real-to-sim.
\begin{table*}[!htbp]
\centering
\caption{Bradley--Terry scores (\%) produced by different
VLM judges on LIBERO.}
\label{tab:bt_scores_libero}
\begin{tabular}{lcccccc}
\toprule
\textbf{VLM}
& $\pi_{0}$
& $\pi_{0.5}$
& OpenVLA
& X-VLA
& NORA
& SmolVLA \\
\midrule
Qwen3-VL-4B      & 20.32 & 20.53 & 11.62 & 17.89 & 15.09 & 14.56 \\
Qwen3-VL-8B      & 20.47 & 20.65 & 11.57 & 18.13 & 14.79 & 14.39 \\
Qwen3-VL-32B     & 20.58 & 20.76 & 11.48 & 18.37 & 14.68 & 14.13 \\
Gemma-3-4B       & 20.35 & 20.74 & 11.65 & 17.55 & 15.08 & 14.63 \\
Gemma-3-12B      & 20.54 & 20.70 & 11.30 & 18.39 & 14.83 & 14.24 \\
Gemma-3-27B      & 20.20 & 20.52 & 11.58 & 18.24 & 14.94 & 14.52 \\
LLaVA-OV-1.5-4B  & 20.62 & 20.80 & 11.34 & 18.03 & 14.79 & 14.43 \\
LLaVA-OV-1.5-8B  & 20.88 & 21.09 & 11.35 & 17.72 & 14.62 & 14.34 \\
\bottomrule
\end{tabular}
\end{table*}

\begin{table*}[!htbp]
\centering
\caption{Bradley--Terry scores (\%) produced by different
VLM judges on Real-to-Sim.}
\label{tab:bt_scores_real_to_sim}
\begin{tabular}{lcccccc}
\toprule
\textbf{VLM}
& $\pi_{0}$
& $\pi_{0.5}$
& OpenVLA
& X-VLA
& NORA
& SmolVLA \\
\midrule
Qwen3-VL-4B      & 24.03 & 34.42 & 9.76 & 25.52 & 6.20 & 0.07 \\
Qwen3-VL-8B      & 24.69 & 31.70 & 9.12 & 29.24 & 5.23 & 0.03 \\
Qwen3-VL-32B     & 24.62 & 28.02 & 8.19 & 34.54 & 4.63 & 0.00 \\
Gemma-3-4B       & 25.37 & 34.50 & 7.89 & 30.00 & 2.13 & 0.11 \\
Gemma-3-12B      & 24.86 & 30.05 & 7.93 & 33.77 & 3.32 & 0.07 \\
Gemma-3-27B      & 24.43 & 31.28 & 9.11 & 28.64 & 6.53 & 0.00 \\
LLaVA-OV-1.5-4B  & 23.22 & 28.36 & 8.70 & 30.50 & 9.12 & 0.10 \\
LLaVA-OV-1.5-8B  & 24.68 & 31.19 & 9.35 & 28.28 & 6.24 & 0.27 \\
\bottomrule
\end{tabular}
\end{table*}

\appsubsection{Human Annotation Protocol}
For each setting, the same three annotators independently evaluate 200
task-stratified video pairs. The identities of the policies are hidden and the \(A/B\) order is randomized independently. Annotators receive the
task instruction and the same videos used for VLM evaluation and select the
video with higher execution quality according to the same criteria in Sec.~\ref{vlm_eval}. The majority vote defines the human reference preference. The annotation set contains 66 success--success, 67 failure--failure, and 67 success--failure pairs in both LIBERO and real-to-sim settings. The unanimous agreement among the annotators is 78.5\% for LIBERO and 80.0\% for real-to-sim. The pairs are distributed uniformly across the 40 LIBERO tasks and as evenly
as possible across the seven real-to-sim tasks. The annotator instructions are as follows. The placeholder \texttt{\{video\_block\}} denotes the input videos.

\begin{lstlisting}[style=prompt]
You are an expert in robotic performance comparison. You must compare Robot A and Robot B based on action details, continuity, fluidity, and task success, and clearly determine which robot performs better.

{video_block}
Your goal is to decide which video is better based ONLY on these criteria:
1. Motion Smoothness
2. Continuity / Fluidity
3. Task Completion
4. Force / Effort Control

Rules:
- Use ONLY the information explicitly stated in the videos.
- Do NOT hallucinate or infer unstated details.
- Smooth, continuous, steady actions → positive evidence.
- Jerky, hesitant, or multiple unnecessary adjustments → negative evidence.

Final Output Rule:
- Output ONLY one uppercase letter: "A" or "B".
- Do NOT output explanations, reasoning, analysis, or any other text.
\end{lstlisting}

\appsubsection{Order-Swap Evaluation}
For each VLM evaluator, we reuse the same video
descriptions and repeat preference evaluation after reversing the \(A/B\)
presentation order for each task-stratified real-to-sim video pair. A comparison is considered consistent when both queries
select the same underlying video after mapping the outputs back to video
identity. The mean order-swap consistency across the eight VLM evaluators is 91.5\%. The swap test is used only as a position-bias audit and does not alter the main preference matrix.

\appsubsection{Pairwise Sampling Details}
For each unordered policy pair, we evaluate checkpoints from
500 to 2,000 comparisons per
unordered policy pair in increments of 100. Comparisons are distributed
as evenly as possible across tasks, with counts differing by at most one when
the budget is not divisible by the number of tasks. The two videos in each
comparison share the same task and initial configuration. Task-level win counts
are pooled into one preference matrix before BT fitting. The video pairs are sampled with replacement. 

\FloatBarrier

\begin{figure}[!htbp]
    \centering
    \includegraphics[width=1.0\textwidth]{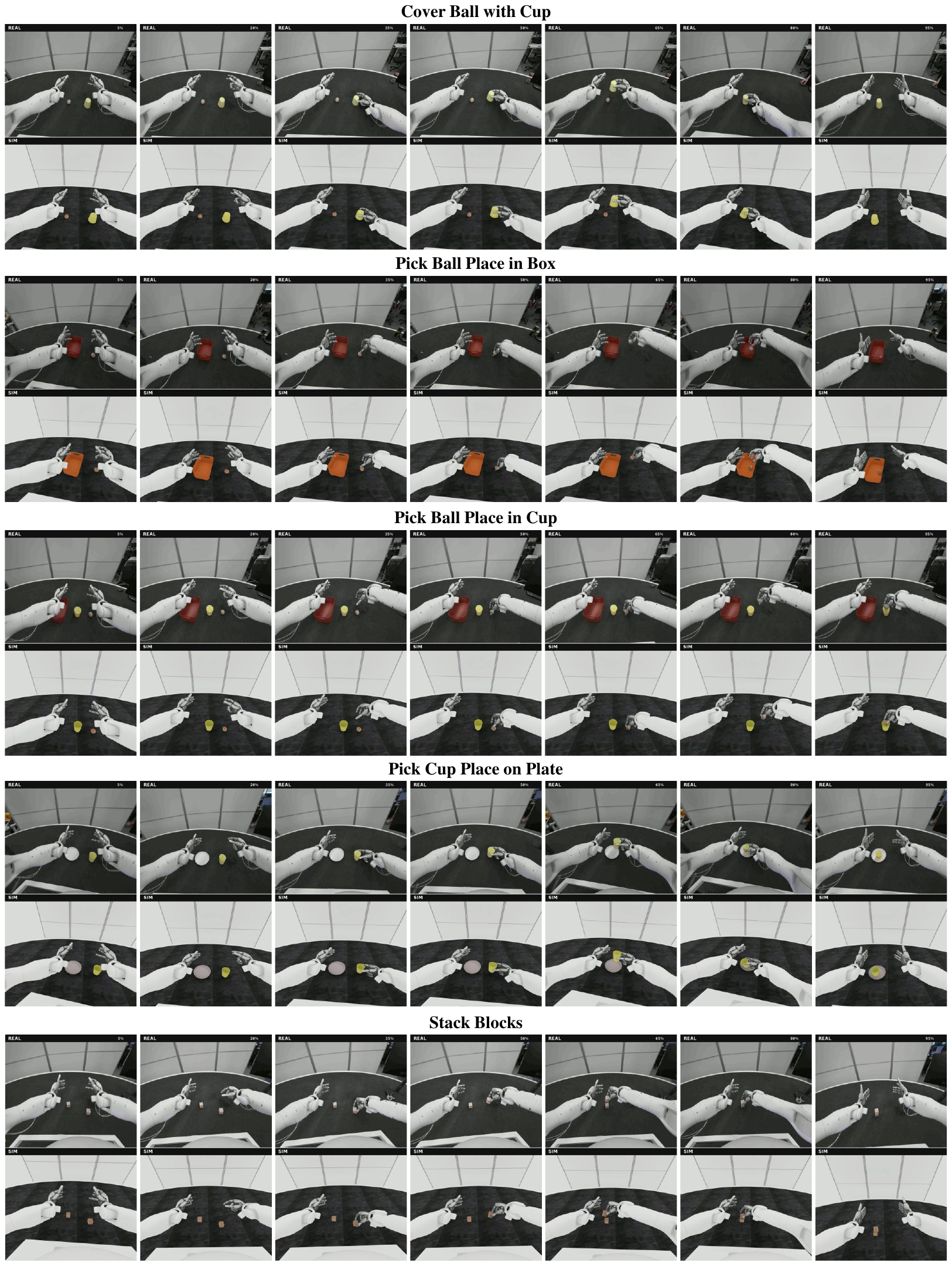}
    \vspace{-2.0em}
    \caption{Examples of real-to-sim trajectory correspondence. For each task, the top row shows the real-world trajectory and the bottom row shows its simulated action-replay counterpart. Each row contains seven frames ordered from left to right.}
    \label{fig:action_replay}
\end{figure}

\begin{figure}[!htbp]
    \centering
    \includegraphics[width=1.0\textwidth]{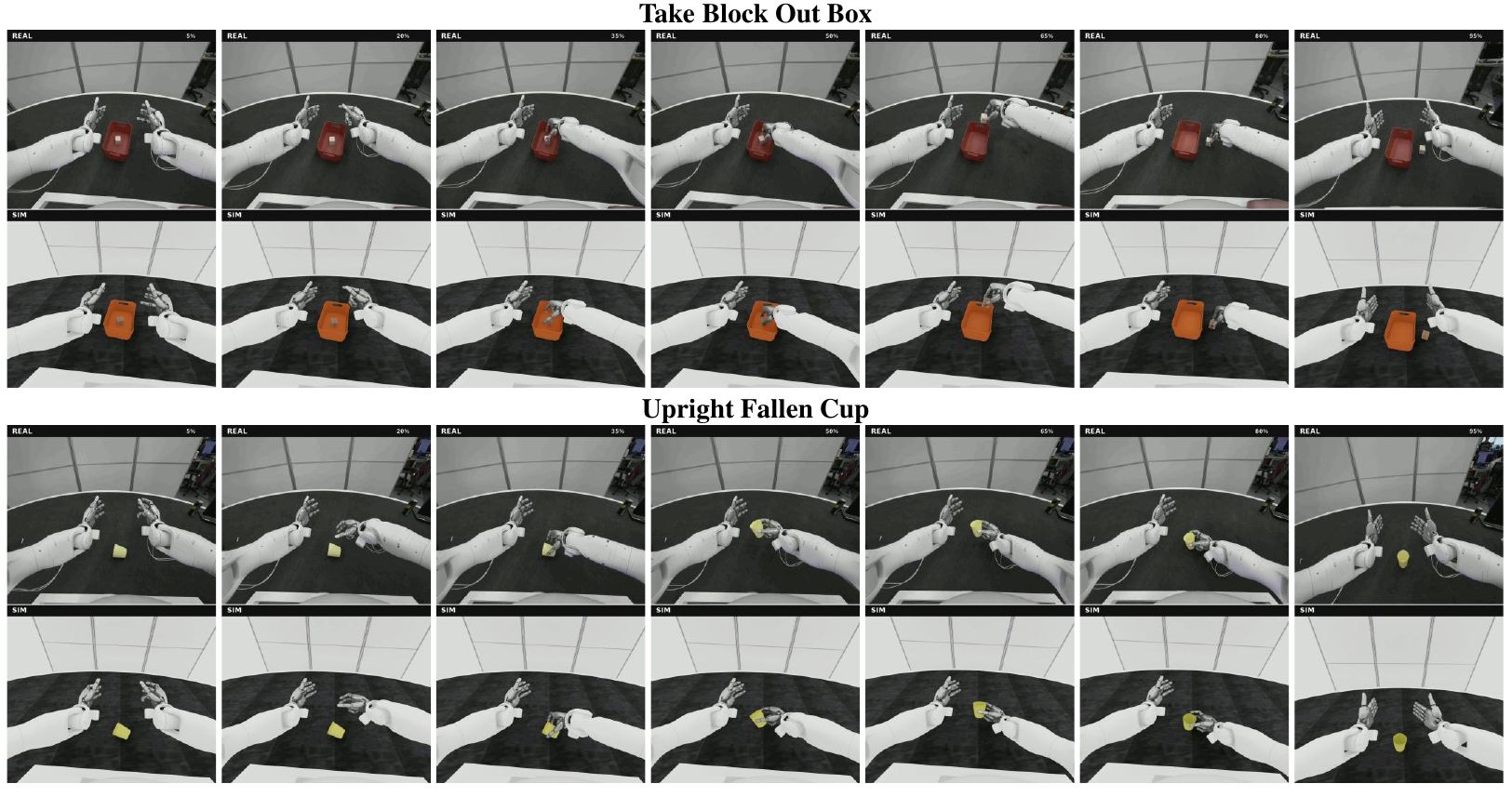}
    \addtocounter{figure}{-1}
    \caption{Examples of real-to-sim trajectory correspondence (continued).}
    \label{fig:action_replay}
\end{figure}

\begin{table}[!htbp]
    \centering

    \begin{minipage}[t]{0.485\linewidth}
        \vspace{0pt}
        \captionof{table}{Training hyperparameters for \(\pi_0\).}
        \label{tab:pi0_hyperparameters}

        \centering
        \footnotesize
        \setlength{\tabcolsep}{3pt}
        \renewcommand{\arraystretch}{1.05}

        \begin{tabularx}{\linewidth}{
            @{}
            >{\raggedright\arraybackslash}p{0.43\linewidth}
            >{\raggedright\arraybackslash}X
            @{}
        }
            \toprule
            Hyperparameter & Value \\
            \midrule
            Fine-tuning method
                & Full fine-tuning (no freeze, no LoRA) \\
            Action representation
                & Flow matching, delta actions \((a-s)\) \\
            Action dimension
                & 65 \\
            Action horizon
                & 25 \\
            Max token length
                & 48 \\
            Cameras / image keys
                & 4 (head L/R stereo, left/right wrist) \\
            Image resolution
                & \(224 \times 224\) \\
            Global batch size
                & 32 \\
            Training steps
                & 300,000 \\
            Optimizer
                & AdamW
                \((\beta_1=0.9,\,
                \beta_2=0.95,\,
                \epsilon=10^{-8})\) \\
            Weight decay
                & \(1\times10^{-10}\) \\
            Gradient clip (global norm)
                & 1.0 \\
            LR schedule
                & Cosine decay \\
            Peak / end LR
                & \(2.5\times10^{-5}/2.5\times10^{-6}\) \\
            Warmup steps
                & 2,000 \\
            EMA decay
                & 0.9999 \\
            \bottomrule
        \end{tabularx}
    \end{minipage}
    \hfill
    \begin{minipage}[t]{0.485\linewidth}
        \vspace{0pt}
        \captionof{table}{Training hyperparameters for \(\pi_{0.5}\).}
        \label{tab:pi05_hyperparameters}

        \centering
        \footnotesize
        \setlength{\tabcolsep}{3pt}
        \renewcommand{\arraystretch}{1.05}

        \begin{tabularx}{\linewidth}{
            @{}
            >{\raggedright\arraybackslash}p{0.43\linewidth}
            >{\raggedright\arraybackslash}X
            @{}
        }
            \toprule
            Hyperparameter & Value \\
            \midrule
            Fine-tuning method
                & Full fine-tuning (no freeze, no LoRA) \\
            Action representation
                & Flow matching, delta actions \((a-s)\) \\
            State input
                & Discrete language tokens \\
            Action dimension
                & 65 \\
            Action horizon
                & 25 \\
            Max token length
                & 300 \\
            Cameras / image keys
                & 4 (head L/R stereo, left/right wrist) \\
            Image resolution
                & \(224 \times224\) \\
            Global batch size
                & 32 \\
            Training steps
                & 300,000 \\
            Optimizer
                & AdamW
                \((\beta_1=0.9,\,
                \beta_2=0.95,\,
                \epsilon=10^{-8})\) \\
            Weight decay
                & \(1\times10^{-10}\) \\
            Gradient clip (global norm)
                & 1.0 \\
            LR schedule
                & Cosine decay \\
            Peak / end LR
                & \(2.5\times10^{-5}/2.5\times10^{-6}\) \\
            Warmup steps
                & 2,000 \\
            EMA decay
                & 0.9999 \\
            \bottomrule
        \end{tabularx}
    \end{minipage}
\end{table}

\begin{table}[!htbp]
    \centering

    \begin{minipage}[t]{0.485\linewidth}
        \vspace{0pt}
        \captionof{table}{Training hyperparameters for OpenVLA.}
        \label{tab:openvla_hyperparameters}

        \centering
        \footnotesize
        \setlength{\tabcolsep}{3pt}
        \renewcommand{\arraystretch}{1.05}

        \begin{tabularx}{\linewidth}{
            @{}
            >{\raggedright\arraybackslash}p{0.43\linewidth}
            >{\raggedright\arraybackslash}X
            @{}
        }
            \toprule
            Hyperparameter & Value \\
            \midrule
            Fine-tuning method            & LoRA (rank 32, $\alpha{=}16$, dropout 0.0) \\
            LoRA target                   & all-linear (Gaussian init) \\
            Action representation         & Discrete tokens (256 bins/dim) \\
            Action dim                    & 65 \\
            Action chunk                  & 25 (current + 24 future) \\
            Action normalization          & q01/q99 bounds $\to[-1,1]$ \\
            Cameras / image keys          & 4 (head L/R stereo, left+right wrist) \\
            Image resolution              & $224\times224$ (no aug) \\
            Per-device / global batch     & 4 / 32 \\
            Grad accumulation             & 1 \\
            Training steps                & 100{,}000 \\
            Optimizer                     & AdamW \\
            Learning rate                 & $5\times10^{-4}$ (constant, no warmup) \\
            Gradient clip                 & None \\
            Shuffle buffer                & 100{,}000 \\
            \bottomrule
        \end{tabularx}
    \end{minipage}
    \hfill
    \begin{minipage}[t]{0.485\linewidth}
        \vspace{0pt}
        \captionof{table}{Training hyperparameters for NORA-LONG.}
        \label{tab:nora_hyperparameters}

        \centering
        \footnotesize
        \setlength{\tabcolsep}{3pt}
        \renewcommand{\arraystretch}{1.05}

        \begin{tabularx}{\linewidth}{
            @{}
            >{\raggedright\arraybackslash}p{0.43\linewidth}
            >{\raggedright\arraybackslash}X
            @{}
        }
            \toprule
            Hyperparameter & Value \\
            \midrule
            Fine-tuning method            & Full fine-tuning (no LoRA, no quantization) \\
            Action representation         & FAST+ tokenizer \\
            Action dim                    & 65 \\
            Action horizon                & 25 (current + 24 future) \\
            Action normalization          & q01/q99 bounds $\to[-1,1]$ \\
            Cameras / image keys          & 4 (head L/R stereo, left+right wrist) \\
            Image resolution              & $224\times224$ \\
            Per-device / global batch     & 8 / 64 \\
            Grad accumulation             & 1 \\
            Training steps                & 200{,}000 \\
            Optimizer                     & AdamW ($\beta_1{=}0.9,\ \beta_2{=}0.95,\ \epsilon{=}10^{-8}$) \\
            Weight decay                  & $1\times10^{-8}$ \\
            Gradient clip                 & 1.0 \\
            LR schedule                   & Cosine \\
            Learning rate                 & $5\times10^{-5}$ \\
            Warmup steps                  & 1{,}000 \\
            Shuffle buffer                & 25{,}000 \\
            \bottomrule
        \end{tabularx}
    \end{minipage}
\end{table}

\FloatBarrier

\begin{table}[!htbp]
    \centering

    \begin{minipage}[t]{0.485\linewidth}
        \vspace{0pt}
        \captionof{table}{Training hyperparameters for X-VLA.}
        \label{tab:xvla_hyperparameters}

        \centering
        \footnotesize
        \setlength{\tabcolsep}{3pt}
        \renewcommand{\arraystretch}{1.05}

        \begin{tabularx}{\linewidth}{
            @{}
            >{\raggedright\arraybackslash}p{0.43\linewidth}
            >{\raggedright\arraybackslash}X
            @{}
        }
            \toprule
            Hyperparameter & Value \\
            \midrule
            Soft prompt                   & 30 domains, len 32; domain\_id = 20 \\
            Action representation         & Flow matching (10 denoise steps at inference) \\
            Action dim                    & 65 \\
            Action chunk                  & 25 \\
            Cameras / image keys          & 4 (head L/R stereo, left+right wrist) \\
            Per-device / global batch     & 16 / 128 \\
            Grad accumulation             & 1 \\
            Training steps                & 200{,}000 \\
            Optimizer                     & AdamW ($\beta_1{=}0.9,\ \beta_2{=}0.95$) \\
            Weight decay                  & 0.0 \\
            Gradient clip                 & 1.0 \\
            Learning rate                 & $5\times10^{-5}$ (learning\_coef 0.1: VLM/prompt $5\times10^{-6}$) \\
            Freeze steps                  & 1{,}000 (only heads+prompts trained first) \\
            Warmup steps                  & 2{,}000 (linear); constant afterwards \\
            \bottomrule
        \end{tabularx}
    \end{minipage}
    \hfill
    \begin{minipage}[t]{0.485\linewidth}
        \vspace{0pt}
        \captionof{table}{Training hyperparameters for SmolVLA.}
        \label{tab:smolvla_hyperparameters}

        \centering
        \footnotesize
        \setlength{\tabcolsep}{3pt}
        \renewcommand{\arraystretch}{1.05}

        \begin{tabularx}{\linewidth}{
            @{}
            >{\raggedright\arraybackslash}p{0.43\linewidth}
            >{\raggedright\arraybackslash}X
            @{}
        }
            \toprule
            Hyperparameter & Value \\
            \midrule
            Fine-tuning method            & Full fine-tuning (vision encoder unfrozen) \\
            Action representation         & Flow matching (10 denoise steps) \\
            Chunk size                    & 25 \\
            Obs steps                     & 1 \\
            Action dim                    & 65 \\
            Cameras / image keys          & 4 (head L/R stereo, left+right wrist) \\
            Image resolution              & $224\times224$ \\
            VLM layers / attention        & 16, cross-attn; expert width $\times0.75$ \\
            Tokenizer max length          & 48 \\
            Per-device / global batch     & 16 / 128 \\
            Grad accumulation             & 1 \\
            Training steps                & 200{,}000 \\
            Optimizer                     & AdamW ($\beta_1{=}0.9,\ \beta_2{=}0.95,\ \epsilon{=}10^{-8}$) \\
            Weight decay                  & $1\times10^{-10}$ \\
            Gradient clip (norm)          & 10 \\
            LR schedule                   & Cosine decay with warmup \\
            Peak / end LR                 & $5\times10^{-5}$ / $2.5\times10^{-6}$ \\
            Warmup / decay steps          & 1{,}000 / 200{,}000 \\
            \bottomrule
        \end{tabularx}
    \end{minipage}
\end{table}

\end{document}